%% file: neurips_2024.tex
\documentclass{article}

\usepackage[preprint]{neurips_2024}

\input{preamble_packages}

\input{preamble_symbols}

\input{shortcuts}

\title{Fast \textsc{Trac}:\\
A Parameter-Free Optimizer for \\Lifelong Reinforcement Learning}

\author{%
  Aneesh Muppidi\\
  Harvard College\\
  \texttt{aneeshmuppidi@college.harvard.edu}\\
  \And
  Zhiyu Zhang\\
  Harvard University\\
  \texttt{zhiyuz@seas.harvard.edu}\\
  \And
  Heng Yang\\
  Harvard University\\
  \texttt{hankyang@seas.harvard.edu}
}

\begin{document}

\maketitle

\begin{abstract}
A key challenge in lifelong reinforcement learning (RL) is the loss of plasticity, where previous learning progress hinders an agent's adaptation to new tasks. While regularization and resetting can help, they require precise hyperparameter selection at the outset and environment-dependent adjustments. Building on the principled theory of online convex optimization, we present a parameter-free optimizer for lifelong RL, called \textsc{Trac}, which requires no tuning or prior knowledge about the distribution shifts. Extensive experiments on Procgen, Atari, and Gym Control environments show that \textsc{Trac} works surprisingly well—mitigating loss of plasticity and rapidly adapting to challenging distribution shifts—despite the underlying optimization problem being nonconvex and nonstationary.
\end{abstract}

\input{sections/introduction}
\input{sections/preliminary}
\input{sections/approach}
\input{sections/experiments}

\input{sections/discussion}
\input{sections/conclusion}
\input{sections/acknowledgments}

\bibliographystyle{rlc}
\bibliography{refs}
\renewcommand*{\bibfont}{\small}

\newpage
\section*{Appendix}
\appendix

\input{sections/appendix}
\end{document}

%% file: preamble_packages.tex
\usepackage{relsize}
\usepackage{ifthen}
\usepackage[colorinlistoftodos]{todonotes}
\usepackage{tabularx}

\usepackage{microtype}
\usepackage{algorithm}
\usepackage{algorithmic}
\usepackage{hyperref}
\hypersetup{
    colorlinks,
    allcolors={blue!50!black}
}
\colorlet{linkequation}{blue!50!black}
\renewcommand*{\bibfont}{\small}

\usepackage{graphics}
\usepackage{colortbl}
\usepackage{xcolor}

\usepackage{rotating}
\usepackage{color}
\usepackage{enumerate}
\usepackage[T1]{fontenc}
\usepackage{booktabs}
\usepackage{graphicx,url}
\usepackage{multirow}
\usepackage{array}
\usepackage{latexsym}
\usepackage{amsfonts}
\usepackage{amsmath}
\usepackage{amssymb}
\usepackage{mathtools}
\usepackage{wrapfig}
\usepackage{xstring}
\usepackage{multirow}
\usepackage{xcolor}
\usepackage{prettyref}
\usepackage{flexisym}
\usepackage{bigdelim}
\usepackage{breqn} 
\usepackage{listings}

\usepackage{enumitem}
\usepackage{xspace}
\usepackage{bm}

%% file: preamble_symbols.tex
\newcommand{\reg}{\mathrm{Regret}}
\newcommand{\defeq}{\mathrel{\mathop:}=}
\newcommand\inner[2]{\left\langle #1, #2 \right\rangle}

\newcommand{\bdmath}{\begin{dmath}}
\newcommand{\edmath}{\end{dmath}}
\newcommand{\beq}{\begin{equation}}
\newcommand{\eeq}{\end{equation}}
\newcommand{\bdm}{\begin{displaymath}}
\newcommand{\edm}{\end{displaymath}}
\newcommand{\bea}{\begin{eqnarray}}
\newcommand{\eea}{\end{eqnarray}}
\newcommand{\beal}{\beq \begin{array}{ll}}
\newcommand{\eeal}{\end{array} \eeq}
\newcommand{\beas}{\begin{eqnarray*}}
\newcommand{\eeas}{\end{eqnarray*}}
\newcommand{\ba}{\begin{array}}
\newcommand{\ea}{\end{array}}
\newcommand{\bit}{\begin{itemize}}
\newcommand{\eit}{\end{itemize}}
\newcommand{\ben}{\begin{enumerate}}
\newcommand{\een}{\end{enumerate}}

\newcommand{\calA}{{\cal A}}

\newcommand{\calG}{{\cal G}}

\newcommand{\calS}{{\cal S}}

\newcommand{\calU}{{\cal U}}

\newcommand{\hiddenText}{{\color{gray} hidden text.}}
\newcommand{\hideWithText}[1]{\hiddenText}

\newcommand{\norm}[1]{\left\| #1 \right\|}

\newcommand{\blue}[1]{{\color{blue}#1}}

\newcommand{\linkToPdf}[1]{\href{#1}{\blue{(pdf)}}}
\newcommand{\linkToPpt}[1]{\href{#1}{\blue{(ppt)}}}
\newcommand{\linkToCode}[1]{\href{#1}{\blue{(code)}}}
\newcommand{\linkToWeb}[1]{\href{#1}{\blue{(web)}}}
\newcommand{\linkToVideo}[1]{\href{#1}{\blue{(video)}}}
\newcommand{\linkToMedia}[1]{\href{#1}{\blue{(media)}}}
\newcommand{\award}[1]{\xspace} 



%% file: shortcuts.tex
\newcommand{\rpar}[1]{\left({#1}\right)}
\newcommand{\spar}[1]{\left[{#1}\right]}
\newcommand{\R}{\mathbb{R}} 

\newcommand{\eps}{\varepsilon} 
\newcommand{\erfi}{\mathrm{erfi}}
\newcommand{\refrm}{\mathrm{ref}}
\newcommand{\base}{\mathrm{Base}}

%% file: sections/introduction.tex

\section{Introduction}
\begin{wrapfigure}{R}{0.47\textwidth}
\vspace{-10pt}
    \centering
    \includegraphics[width=0.45\textwidth]{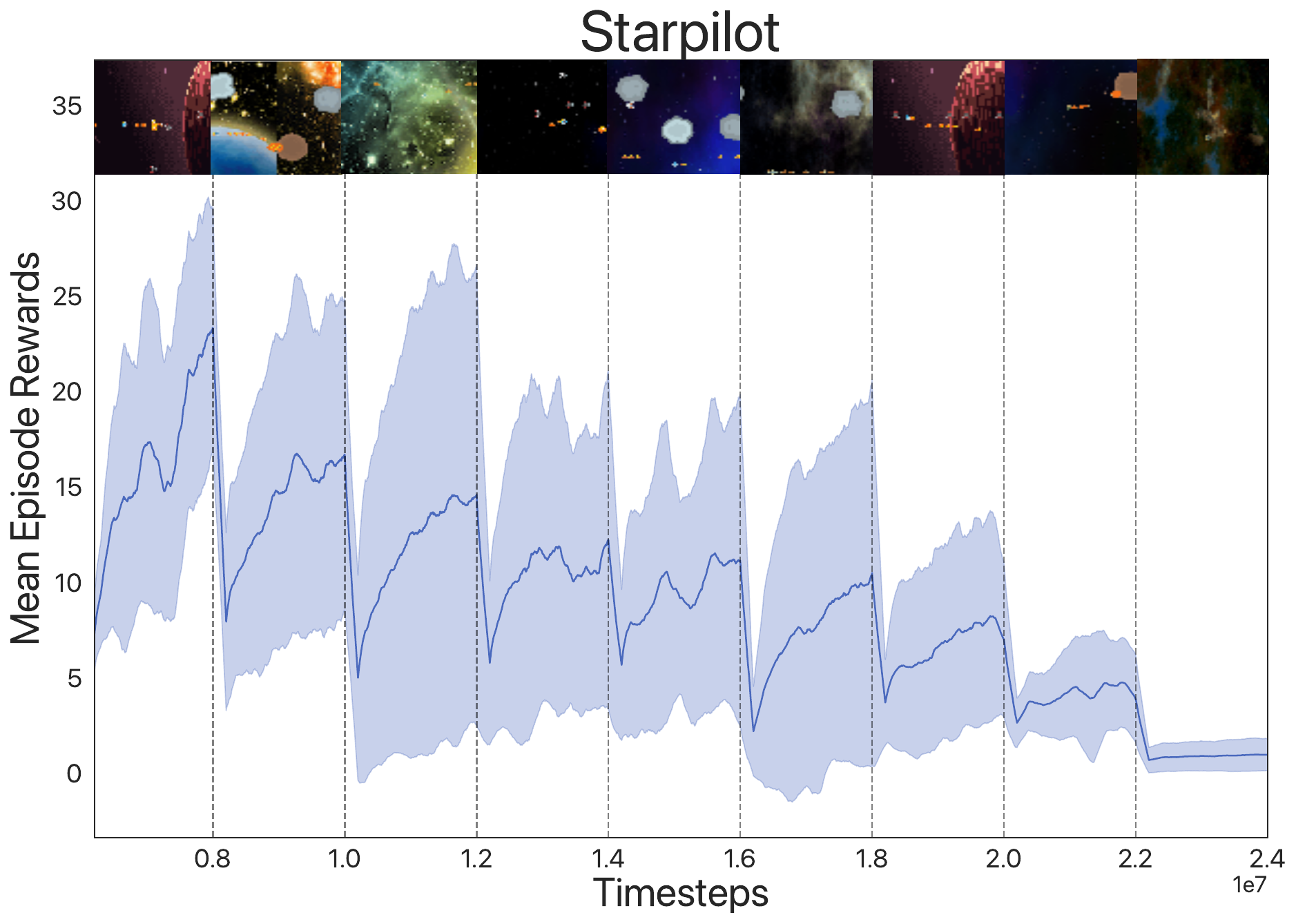}
    \caption{Severe loss of plasticity in Procgen (Starpilot). There is a steady decline in reward with each distribution shift.}
    \vspace{-6pt}
    \label{fig:loss_of_plasticity}
\end{wrapfigure}

Spot, the agile robot dog, has been learning to walk confidently across soft, lush grass. But when Spot moves from the grassy field to a gravel surface, the small stones shift beneath her feet, causing her to stumble. When Spot tries to walk across a sandy beach or on ice, the challenges multiply, and her once-steady walk becomes erratic. Spot wants to adjust quickly to these new terrains, but the patterns she learned on grass are not suited to gravel, sand, or ice. Furthermore, she never knows when the terrain will change again and how different it will be, therefore must continually plan for the unknown while avoiding reliance on outdated experiences.

Spot's struggle exemplifies a well-known and extensively studied challenge in real-world decision making: \emph{lifelong reinforcement learning} (lifelong RL) \cite{abel2023definition, ben2023lifelong, mendez2020lifelong, xie2022lifelong}. In lifelong RL, the learning agent must continually acquire new knowledge to adapt to the nonstationarity of the environment. At first glance, there appears to be an obvious solution: given a policy gradient oracle, the agent could just keep running gradient descent nonstop. However, recent experiments have demonstrated an intriguing behavior called \emph{loss of plasticity} \citep{dohare2021continual, lyle2022understanding, pmlr-v232-abbas23a, sokar2023dormant}: despite persistent gradient steps, such an agent can gradually lose its responsiveness to incoming observations. There are even extreme cases of loss of plasticity (known as \emph{negative transfer} or \emph{primacy bias}), where prior learning can significantly hamper the performance in new tasks \citep{nikishin2022primacy, ahn2024catastrophic}; see Figure~\ref{fig:loss_of_plasticity} for an example. All these suggest that the problem is more involved than one might think.

From the optimization perspective, the above issues might be attributed to the \emph{lack of stability} under gradient descent. That is, the weights of the agent's parameterized policy can drift far away from the origin (or a good initialization), leading to a variety of undesirable behaviors.\footnote{Such as the inactivation of many neurons, due to the ReLU activation function \citep{pmlr-v232-abbas23a,sokar2023dormant}.} Fitting this narrative, it has been shown that simply adding a $L_2$ regularizer to the optimization objective \citep{kumar2023maintaining} or periodically resetting the weights \citep{dohare2021continual,asadi2023resetting,sokar2023dormant,ahn2024catastrophic} can help mitigate the problem. However, a particularly important limitation is their use of \emph{hyperparameters}, such as the magnitude of the regularizer and the resetting frequency\footnote{Indeed, hyperparameter selection, in general, is a well-known problem in lifelong as well as continual learning settings \citep{de2021continual}.}. Good performance hinges on the suitable environment-dependent hyperparameter, but how can one confidently choose that \emph{before} interacting with the environment?  The classical cross-validation approach would violate the one-shot nature of lifelong RL (and online learning in general; see Chapter~1 of \citealp{orabona2023modern}), since it is impossible to experience the same environment multiple times. This leads to the contributions of the present work.

\paragraph{Contribution} The present work addresses the key challenges in lifelong RL using the principled theory of \emph{Online Convex Optimization} (OCO). Specifically, our contributions are two fold. 
\begin{itemize}[topsep=0pt,leftmargin=*]
\item \textbf{Algorithm: \textsc{Trac}}\quad Building on a series of results in OCO \citep{cutkosky2018black,cutkosky2019combining,cutkosky2023mechanic,zhang2024improving}, we propose a (hyper)-\emph{parameter-free} optimizer for lifelong RL, called \textsc{\textbf{Trac}} (Adap\textbf{T}ive \textbf{R}egulariz\textbf{A}tion in \textbf{C}ontinual environments). Intuitively, the idea is a refinement of regularization: instead of manually selecting the magnitude of regularization beforehand, \textsc{Trac} chooses that in an online, data-dependent manner. From the perspective of OCO theory, \textsc{Trac} is insensitive to its own hyperparameter, which means that no hyperparameter tuning is necessary in practice. Furthermore, as an optimization approach to lifelong RL, \textsc{Trac} is compatible with any policy parameterization method. 

\item \textbf{Experiment}\quad Using \emph{Proximal Policy Optimization} (PPO) \citep{schulman2017proximal}, we conduct comprehensive experiments on the instantiation of \textsc{Trac} called \textsc{Trac} PPO. A diverse range of lifelong RL environments are tested (based on Procgen, Atari, and Gym Control), with considerably larger scale than prior works.  In settings where existing approaches \citep{pmlr-v232-abbas23a,kumar2023maintaining, ben2023lifelong} struggle, we find that \textsc{Trac} PPO 
 \begin{itemize}[topsep=0pt,leftmargin=*]
 \item mitigates mild and extreme loss of plasticity; 
 \item and rapidly adapts to new tasks when distribution shifts are introduced.
 \end{itemize}
  Such findings might be surprising: the theoretical advantage of \textsc{Trac} is motivated by the convexity in OCO, but lifelong RL is \emph{both nonconvex and nonstationary} in terms of optimization. 
\end{itemize}

\paragraph{Organization} Section~\ref{section:setting} surveys the basics of lifelong RL. Section~\ref{section:method} introduces our parameter-free algorithm \textsc{Trac}, and experiments are presented in Section~\ref{section:experiment}. We defer the discussion of related works and results to Section~\ref{section:discussion}. Finally, Section~\ref{section:conclusion} concludes the paper. 

%% file: sections/preliminary.tex

\section{Lifelong RL}\label{section:setting}

As a sequential decision making framework, \emph{reinforcement learning} (RL) is commonly framed as a \emph{Markov Decision Process} (MDP) defined by the state space $\calS$, the action space $\calA$, the transition dynamics $P(s_{t+1}|s_t,a_t)$, and the reward function $R(s_t,a_t,s_{t+1})$. In the $t$-th round, starting from a state $s_t\in\calS$, the learning agent needs to choose an action $a_t\in\calA$ without knowing $P$ and $R$. Then, the environment samples a new state $s_{t+1}\sim P(\cdot|s_t,a_t)$, and the agent receives a \emph{reward} $r_t=R(s_t,a_t,s_{t+1})$. There are standard MDP objectives driven by theoretical tractability, but from a practical perspective, we measure the agent's performance by its cumulative reward $\sum_{t=1}^Tr_t$.

The standard setting above concerns a \emph{stationary} MDP. Motivated by the prevalence of distribution shifts in practice, the present work studies a nonstationary variant called \emph{lifelong} RL, where the transition dynamics $P_t$ and the reward function $R_t$ can vary over time. Certainly, one should not expect any meaningful ``learning'' against \emph{arbitrary} unstructured nonstationarity. Therefore, we implicitly assume $P_t$ and $R_t$ to be \emph{piecewise constant} over time, and each piece is called a \emph{task} -- just like our example of Spot in the introduction. The main challenge here is to transfer previous learning progress to new tasks. This is reasonable when tasks are similar, but we also want to reduce the degradation when tasks turn out to be very different.

\paragraph{Lifelong RL as online optimization} Deep RL approaches, including PPO \citep{schulman2017proximal} and others, crucially utilize the idea of \emph{policy parameterization}. Specifically, a policy refers to the distribution of the agent's action $a_t$ (conditioned on the historical observations), and we use $\theta_t\in\mathbb{R}^d$ to denote the parameterizing \emph{weight vector}. After sampling $a_t$ and receiving new observations, the agent could define a \emph{loss function} $J_t(\theta)$ that characterizes the ``hypothetical performance'' of each weight $\theta\in\mathbb{R}^d$. Then, by computing the \emph{policy gradient} $g_t=\nabla J_t(\theta_t)$, one could apply a \emph{first order optimization algorithm}\footnote{Formally, a dynamical system that given its state $\theta_t$ and input $g_t$ outputs the new state $\textsc{OPT}(\theta_t,g_t)$.} \textsc{OPT} to obtain the updated weight, $\theta_{t+1}=\textsc{OPT}(\theta_t,g_t)$. 

For the rest of this paper, we will work with such an abstraction. The feedback of the environment is treated as a \emph{policy gradient oracle} $\calG$, which maps the time $t$ and the current weight $\theta_t$ into a policy gradient $g_t=\calG(t,\theta_t)$. Our goal is to design an optimizer \textsc{OPT} well suited for lifelong RL.

\paragraph{Lifelong vs. Continual}
In the RL literature, the use of ``lifelong'' and ``continual'' varies significantly across studies, which may lead to confusion. \cite{abel2023definition} characterized \emph{continual reinforcement learning} (CRL) as a never-ending learning process. However, much of the literature cited under CRL, such as \citep{pmlr-v232-abbas23a,ahn2024catastrophic}, primarily focuses on the problem of \emph{backward transfer} (avoiding catastrophic forgetting). Various policy-based architectures, such as those proposed by \cite{rolnick2019experience, schwarz2018progress, ben2023lifelong}, focus on tackling this issue.  Conversely, the present work addresses the problem of \emph{forward transfer}, which refers to the rapid adaptation to new tasks. Because of this we use ``lifelong'' rather than ``continual'' in our exposition, similar to \citep{Thrun96b,pmlr-v80-abel18b,DBLP:journals/corr/abs-2004-10190}.

%% file: sections/approach.tex

\section{Method}\label{section:method}
\begin{wrapfigure}{R}{0.45\textwidth}
    \centering
    \includegraphics[width=0.44\textwidth]{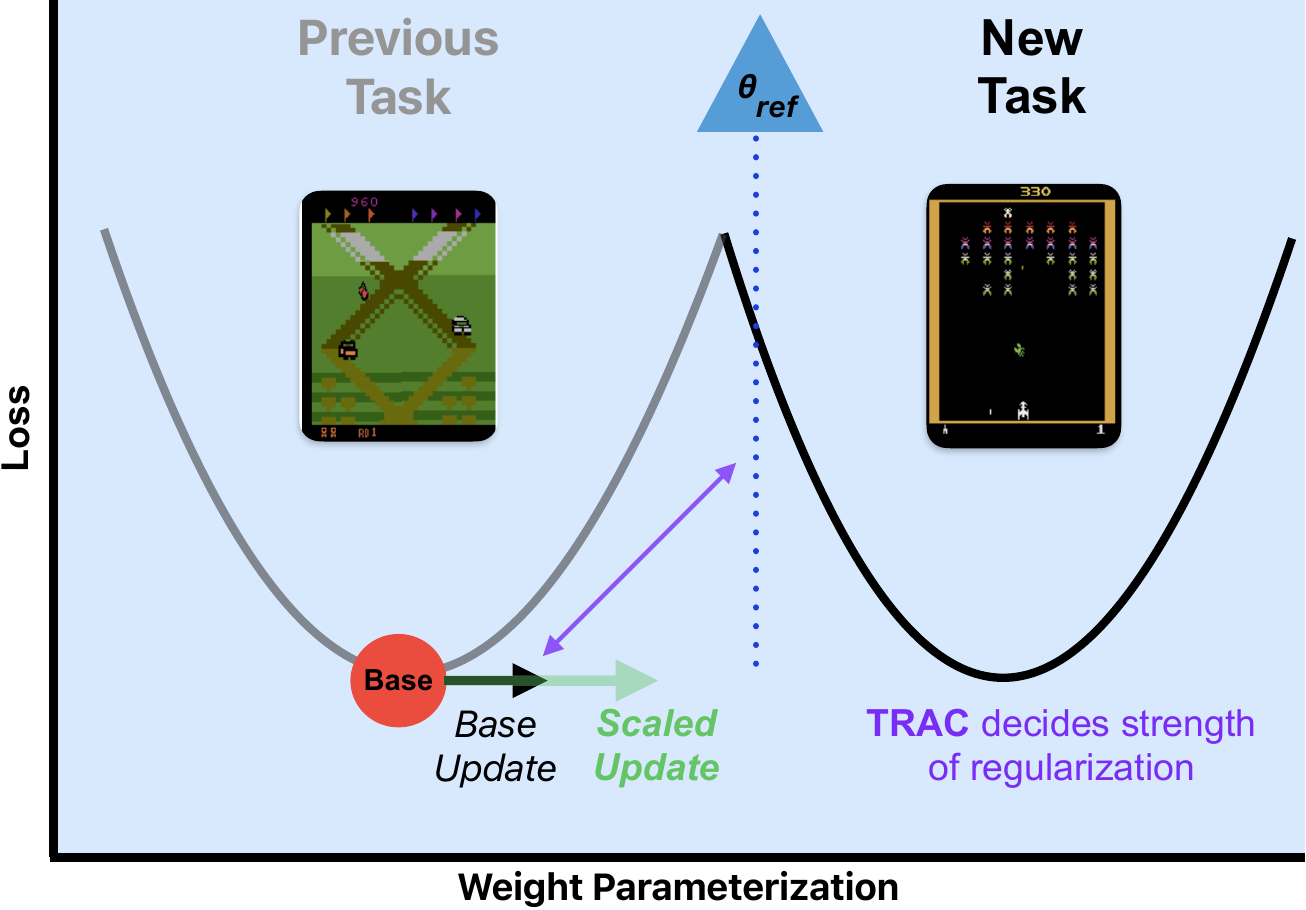}
    \caption{Visualization of \textsc{Trac}'s key idea.}
    \vspace{-6pt}
    \label{fig:concept}
\end{wrapfigure}
Inspired by \citep{cutkosky2023mechanic}, we study lifelong RL by exploiting its connection to \emph{Online Convex Optimization} (OCO; \citealp{zinkevich2003online}). The latter is a classical theoretical problem in online learning, and much effort has been devoted to designing \emph{parameter-free} algorithms that require minimum tuning or prior knowledge \citep{streeter2012no,mcmahan2014unconstrained,orabona2016coin,foster2017parameter,cutkosky2018black,mhammedi2020lipschitz,chen2021impossible,jacobsen2022parameter}. The surprising observation of \cite{cutkosky2023mechanic} is that several algorithmic ideas closely tied to the convexity of OCO can actually improve the nonconvex deep learning training, suggesting certain notions of ``near convexity'' on its loss landscape. We find that lifelong RL (which is \emph{both nonconvex and nonstationary} in terms of optimization) exhibits a similar behavior, therefore a particularly strong algorithm (named \textsc{Trac}) can be obtained from principled results in parameter-free OCO. Let us start from the background.

\paragraph{Basics of (parameter-free) OCO} As a standalone theoretical topic, OCO concerns a sequential optimization problem where the convex loss function $l_t$ can vary arbitrarily over time. In the $t$-th iteration, the optimization algorithm picks an iterate $x_t$ and then observes a gradient $g_t=\nabla l_t(x_t)$. Motivated by the pursuit of ``convergence'' in optimization, the standard objective is to guarantee low (i.e., sublinear in $T$) \emph{static regret}, defined as
\begin{equation*}
\reg_T(l_{1:T},u)\defeq\sum_{t=1}^Tl_t(x_t)-\sum_{t=1}^Tl_t(u),
\end{equation*}
where $T$ is the total number of rounds, and $u$ is a \emph{comparator} that the algorithm does not know beforehand. In other words, the goal is to make $\reg_T(l_{1:T},u)$ small for \emph{all} possible loss sequence $l_{1:T}$ and comparator $u$. Note that for \emph{nonstationary} OCO problems analogous to lifelong RL, it is better to consider a different objective called the \emph{discounted regret}. Algorithms there mostly follow the same principle as in the stationary setting, just wrapped by \emph{loss rescaling} \citep{zhang2024discountedadaptiveonlinelearning}. 

For minimizing static regret, classical \emph{minimax} algorithms like gradient descent \citep{zinkevich2003online} would assume a small \emph{uncertainty set} $\calU$ at the beginning. Then, by setting the hyperparameter (such as the learning rate) according to $\calU$, it is possible to guarantee sublinear \emph{worst case regret},
\begin{equation}\label{eq:minimax_bound}
\max_{(l_{1:T},u)\in\calU}\reg_T(l_{1:T},u)=o(T).
\end{equation}
In contrast, parameter-free algorithms use very different strategies\footnote{The key difference with gradient descent is the use of intricate (non-$L_2$) regularizers. See \citep{fang2022online,jacobsen2022parameter} for a theoretical justification of their importance.} to bound $\reg_T(l_{1:T},u)$ directly (without taking the maximum) by a function of both $l_{1:T}$ and $u$. 
The resulting bound is more refined than Eq.(\ref{eq:minimax_bound}) \citep[Chapter 9]{orabona2023modern}, and crucially, since there is no need to pick an uncertainty set $\calU$, much less hyperparameter tuning is needed. This is where its name comes from.

\textbf{\textsc{TRAC} for Lifelong RL:} In lifelong RL, a key issue is the excessive drifting of weights $\theta_t$, which can detrimentally affect adapting to new tasks. To address this, \textsc{TRAC} enforces proximity to a well-chosen reference point $\theta_{\mathrm{ref}}$, providing a principled solution derived from a decade of research in parameter-free OCO. Unlike traditional methods such as $L_2$ regularization or resetting, \textsc{TRAC} avoids hyperparameter tuning, utilizing the properties of OCO to maintain weight stability and manage the drift effectively.

The core of \textsc{TRAC}, similar to other parameter-free optimizers, incorporates three techniques:
\begin{itemize}
    \item \textbf{Direction-Magnitude Decomposition}: Inspired by \cite{cutkosky2018black}, this technique employs a carefully designed one-dimensional algorithm, the "parameter-free tuner," atop a base optimizer. This setup acts as a data-dependent regularizer, controlling the extent to which the iterates deviate from their initialization, thereby minimizing loss of plasticity, which is crucial given the high plasticity at the initial policy parameterization \citep{pmlr-v232-abbas23a}.
    \item \textbf{Erfi Potential Function}: Building on the previous concept, the tuner utilizes the Erfi potential function, as developed by \cite{zhang2024discountedadaptiveonlinelearning}. This function is crafted to effectively balance the distance of the iterates from both the origin and the empirical optimum. It manages the update magnitude by focusing on the gradient projection along the direction $\theta_t - \theta_{\mathrm{ref}}$.
    \item \textbf{Additive Aggregation:} The tuner above necessitates discounting. Thus, we employ Additive Aggregation by \cite{cutkosky2019combining}. This approach enables the combination of multiple parameter-free OCO algorithms, each with different discount factors, to approximate the performance of the best-performing algorithm. Importantly, it facilitates the automatic selection of the optimal discount factor during training.
\end{itemize}
 
These three components crucially work together to guarantee good regret bounds in the convex setting and are the minimum requirement for any reasonable parameter-free optimizer.

\begin{algorithm*}[ht]
\caption{\textsc{Trac}: Parameter-free Adaption for Continual Environments.\label{alg:meta}}
\begin{algorithmic}[1]
\STATE \textbf{Input:} A policy gradient oracle $\calG$; a first order optimization algorithm \textsc{Base}; a reference point $\theta_{\refrm}\in\R^d$; $n$ discount factors $\beta_1,\ldots,\beta_n\in(0,1]$ (default: $0.9,0.99,\ldots,0.999999$). 
\STATE \textbf{Initialize:} Create $n$ copies of Algorithm~\ref{alg:tuner}, denoted as $\calA_1,\ldots,\calA_n$. For each $j\in[1:n]$, $\calA_j$ uses the discount factor $\beta_j$. Initialize the algorithm \textsc{Base} at $\theta_{\refrm}$. Let $\theta_1=\theta_\refrm$.
\FOR{$t=1,2,\ldots$}
\STATE Obtain the $t$-th policy gradient $g_t=\calG(t,\theta_t)\in\R^d$.
\STATE Send $g_t$ to \textsc{Base} as its $t$-th input, and get its output $\theta_{t+1}^{\base}\in\R^d$.
\STATE For all $j\in[1:n]$, send $\inner{g_t}{\theta_t-\theta_{\refrm}}$ to $\calA_j$ as its $t$-th input, and get its output $s_{t+1,j}\in\R$. 
\STATE Define the scaling parameter $S_{t+1}=\sum_{j=1}^ns_{t+1,j}$.
\STATE Update the weight of the policy,
\begin{equation*}
\theta_{t+1}=\theta_{\refrm}+\rpar{\theta^{\base}_{t+1}-\theta_{\refrm}}S_{t+1}.
\end{equation*}
\ENDFOR
\end{algorithmic}
\end{algorithm*}

\begin{algorithm*}[ht]
\caption{1D Discounted Tuner of \textsc{Trac}.\label{alg:tuner}}
\begin{algorithmic}[1]
\STATE \textbf{Input:} Discount factor $\beta\in(0,1]$; small value $\eps>0$ (default: $10^{-8}$).
\STATE \textbf{Initialize:} The running variance $v_0=0$; the running (negative) sum $\sigma_0=0$.
\FOR{$t=1,2,\ldots$}
\STATE Obtain the $t$-th input $h_t$. 
\STATE Let $v_t=\beta^2v_{t-1}+h^2_t$, and $\sigma_t=\beta \sigma_{t-1}-h_t$.
\STATE Select the $t$-th output
\begin{equation*}
s_{t+1}=\frac{\eps}{\erfi(1/\sqrt{2})}\erfi\rpar{\frac{\sigma_{t}}{\sqrt{2v_{t}}+\eps}},
\end{equation*}
where $\erfi$ is the \emph{imaginary error function} queried from standard software packages. 
\ENDFOR
\end{algorithmic}
\end{algorithm*}

Without going deep into the theory, here is an overview of the important ideas (also see Figure~\ref{fig:concept} for a visualization).
\begin{itemize}[topsep=0pt,leftmargin=*]
\item First, \textsc{Trac} is a meta-algorithm that operates on top of a ``default'' optimizer \textsc{Base}. It can simply be gradient descent with a constant learning rate, or \textsc{Adam} \citep{kingma2014adam} as in our experiments. Applying \textsc{Base} alone would be equivalent to enforcing the scaling parameter $S_{t+1}\equiv 1$ in \textsc{Trac}, but this would suffer from the drifting of $\theta_{t+1}^\base$ (and thus, the weight $\theta_{t+1}$). 
\item To fix this issue, \textsc{Trac} uses the tuner (Algorithm~\ref{alg:tuner}) to select the scaling parameter $S_{t+1}$, making it \emph{data-dependent}. Typically $S_{t+1}$ is within $[0,1]$ (see Figure~\ref{fig:scaling-procgen} to \ref{fig:scaling-control}), therefore essentially, we define the updated weight $\theta_{t+1}$ as a \emph{convex combination} of the \textsc{Base}'s weight $\theta_t^\base$ and the reference point $\theta_\refrm$,
\begin{equation*}
\theta_{t+1}=S_{t+1}\cdot \theta_{t+1}^\base+(1-S_{t+1})\theta_\refrm.
\end{equation*}
This brings the weight closer to $\theta_\refrm$, which is known to be ``safe'' (i.e., not overfitting any particular lifelong RL task), although possibly conservative. 
\item To inject the right amount of conservatism without hyperparameter tuning, the tuner (Algorithm~\ref{alg:tuner}) applies an unusual decision rule based on the $\erfi$ function. Theoretically, this is known to be optimal in an idealized variant of OCO \citep{zhang2022pde,zhang2024improving}, but removing the idealized assumptions requires a tiny bit of extra conservatism, which is challenging (and not necessarily practical). Focusing on the lifelong RL problem that considerably deviates from OCO, we simply apply the $\erfi$ decision rule as is. This is loosely motivated by deep learning training dynamics, e.g., \citep{cohen2020gradient,ahn2023learning,andriushchenko2023sgd}, where an aggressive optimizer is often observed to be better. 
\item Finally, the tuner requires a discount factor $\beta$. This crucially controls the strength of regularization (elaborated next), but also introduces a hyperparameter tuning problem. Following \citep{cutkosky2019combining}, we aggregate tuners with different $\beta$ (on a log-scaled grid) by simply summing up their outputs. This is justified by the \emph{adaptivity} of the tuner itself: in OCO, if we add a parameter-free algorithm $\calA_1$ to any other algorithm $\calA_2$ that already works well, then $\calA_1$ can automatically identify this and ``tune down'' its aggressiveness, such that $\calA_1+\calA_2$ still performs as well as $\calA_2$. 
\end{itemize}

\paragraph{Connection to regularization}
Despite its nested structure, \textsc{Trac} can actually be seen as a parameter-free refinement of $L_2$ regularization \citep{kumar2023maintaining}. To concretely explain this intuition, let us consider the following two optimization dynamics.
\begin{itemize}[topsep=0pt,leftmargin=*]
\item First, suppose we run gradient descent with learning rate $\eta$, on the policy gradient sequence $\{g_t\}$ with the $L_2$ regularizer $\frac{\lambda}{2}\norm{\theta-\theta_\refrm}^2$. Quantitatively, it means that starting from the $t$-th weight $\theta_t$,
\begin{equation}\label{eq:case1}
\theta_{t+1}=\theta_t-\eta\spar{g_t+\lambda\rpar{\theta_t-\theta_\refrm}},\quad\Longrightarrow\quad \theta_{t+1}-\theta_{\refrm}=\rpar{1-\lambda \eta}\rpar{\theta_t-\theta_\refrm}-\eta g_t.
\end{equation}
That is, the updated weight $\theta_{t+1}$ is determined by a ($1-\lambda\eta$)-discounting with respect to the reference point $\theta_\refrm$, followed by a gradient step $-\eta g_t$. 
\item Alternatively, consider applying the following simplification of \textsc{Trac} on the same policy gradient sequence $\{g_t\}$: ($i$) \textsc{Base} is still gradient descent with learning rate $\eta$; ($ii$) there is just one discount factor $\beta$; and ($iii$) the one-dimensional tuner (Algorithm~\ref{alg:tuner}) is replaced by the $\beta$-discounted gradient descent with learning rate $\alpha$, i.e., $S_{t+1}=\beta S_t-\alpha h_t$. In this case, we have
\begin{align*}
\theta_{t+1}-\theta_\refrm&=S_{t+1}\rpar{\theta^\base_{t+1}-\theta_\refrm}\\
&=\rpar{\beta S_t-\alpha h_t}\rpar{\theta^\base_{t}-\theta_\refrm-\eta g_t}\\
&=\rpar{\beta-\alpha S^{-1}_t h_t}\rpar{\theta_{t}-\theta_\refrm}-\eta S_{t+1} g_t.\tag{mildly assuming $S_t\neq 0$}
\end{align*}
Notice that $S_t$ is a $\beta$-discounted sum of $\alpha h_1,\ldots,\alpha h_{t-1}$, thus in the typical situation of $\beta\approx 1$ one might expect $\alpha h_t\ll |S_t|$. Then, the resulting update of $\theta_{t+1}$ is similar to Eq.\eqref{eq:case1}, with quantitative changes on the ``effective discounting'' $1-\lambda\eta\rightarrow\beta$, and the ``effective learning rate'' $\eta\rightarrow \eta S_{t+1}$. 
\end{itemize}

The main message here is that under a simplified setting, \textsc{Trac} is almost equivalent to $L_2$ regularization. The latter requires choosing the hyperparameters $\lambda$ and $\eta$, and similarly, the above \emph{simplified} \textsc{Trac} requires choosing $\beta$ and $\eta$. Going beyond this simplification, the actual \textsc{Trac} removes the tuning of $\beta$ using aggregation, and the tuning of $\eta$ using the $\erfi$ decision rule. 

\paragraph{On the hyperparameters} Although \textsc{Trac} is called ``parameter-free'', it still needs the $\beta$-grid, the constant $\eps$ and the algorithm \textsc{Base} as inputs. The idea is that \textsc{Trac} is particularly insensitive to such choices, as supported by the OCO theory. As the result, the generic default values recommended by \cite{cutkosky2023mechanic} are sufficient in practice. We note that those are proposed for training supervised deep learning models, thus should be agnostic to the lifelong RL applications we consider. 

%% file: sections/experiments.tex

\section{Experiment}\label{section:experiment}

\begin{figure}[h]
    \centering
    \includegraphics[width=1\textwidth]{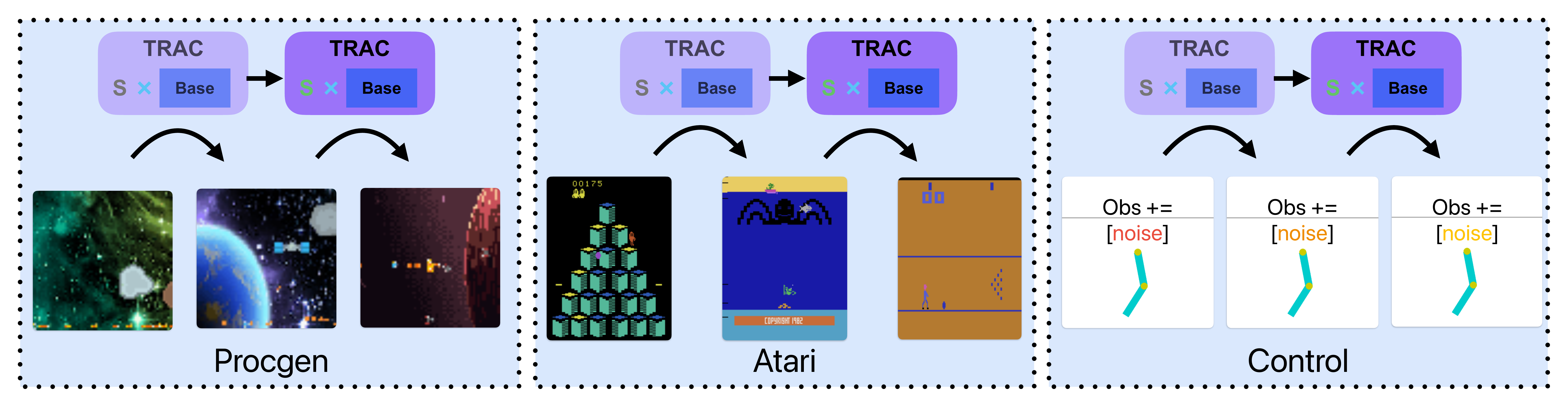}
    \caption{Experimental setup for lifelong RL.}
    \label{fig:exp-setup}
\end{figure}

Does \textsc{Trac} experience the common pitfalls of loss of plasticity? Does it rapidly adapt to distribution shifts? To answer these questions, we test \textsc{Trac} in empirical RL benchmarks such as vision-based games and physics-based control environments in lifelong settings (Figure~\ref{fig:exp-setup}). Specifically, we instantiate PPO with two different optimizers: \textsc{Adam} with constant learning rate for baseline comparison, and \textsc{Trac} for our proposed method (with exactly the same \textsc{Adam} as the input \textsc{Base}). We also test \textsc{Adam} PPO with \emph{concatenated ReLU activations} (CReLU; \citealp{shang2016understanding}), previously shown to mitigate loss of plasticity in certain deep RL settings \citep{pmlr-v232-abbas23a}. Our numerical results are summarized in Table~\ref{tab:cum_reward}. Across every lifelong RL setting, we observe substantial improvements in the cumulative episode reward by using \textsc{Trac} PPO compared to \textsc{Adam} PPO or CReLU. Below are the details, with more in the Appendix.

\textbf{Procgen}\quad We first evaluate on OpenAI Procgen, a suite of 16 procedurally generated game environments \citep{DBLP:journals/corr/abs-1912-01588}. We introduce distribution shifts by sampling a new procedurally generated level of the current game every 2 million time steps, treating each level as a distinct task.

\begin{figure}[ht]
    \centering
    \includegraphics[width=1\textwidth]{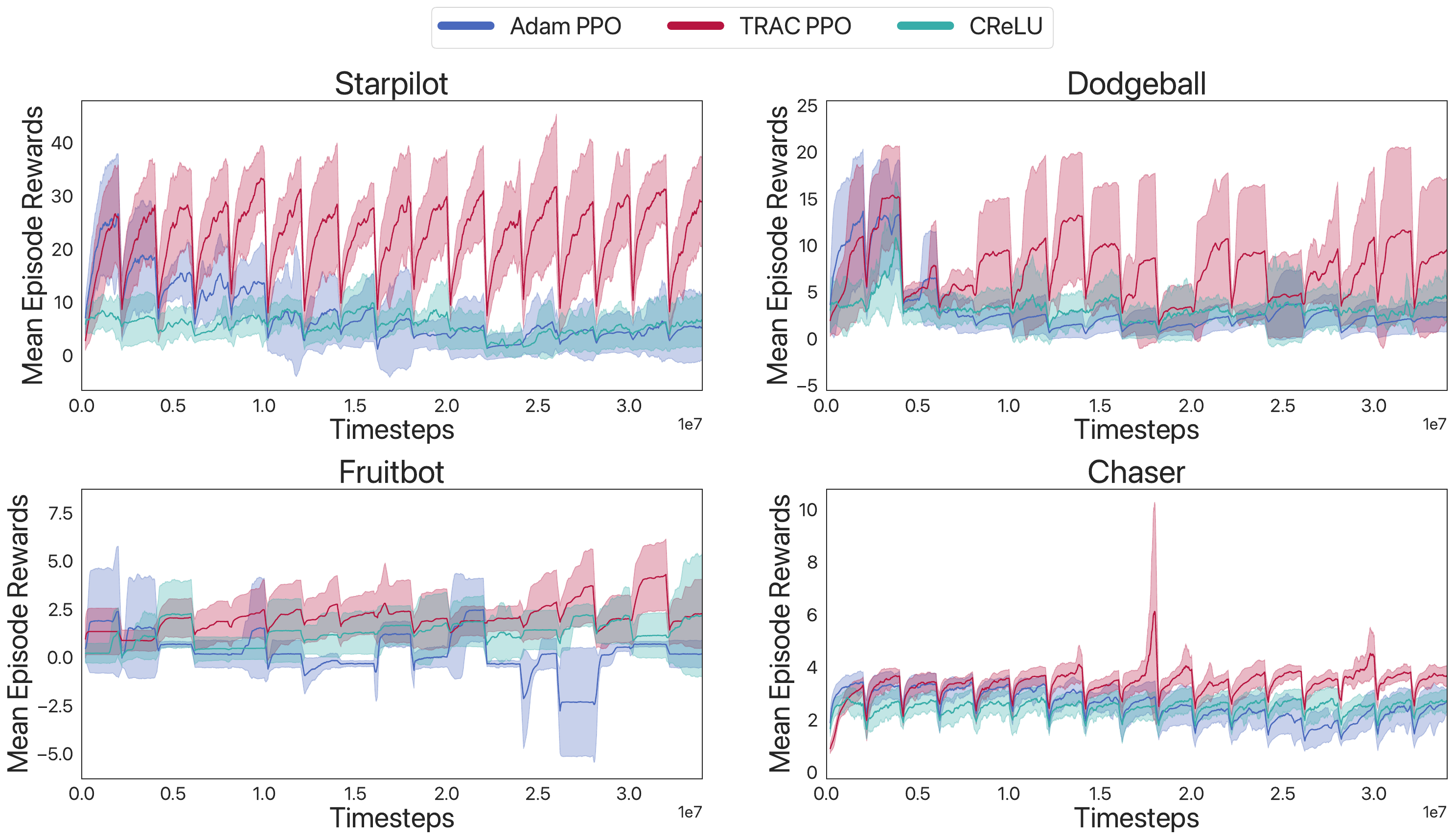}
    \caption{Reward in the lifelong Procgen environments for StarPilot, Dodgeball, Fruitbot, and Chaser. There is a steady loss of plasticity in agents using \textsc{Adam} PPO and CReLU, characterized by their inability to maintain performance through succesive Procgen levels. In contrast, \textsc{Trac} avoids this loss of plasticity, quickly achieving high performance with each new task.}
    \label{fig:procgen}
\end{figure}

We evaluate game environments including StarPilot, Dodgeball, Fruitbot, and Chaser. In all of these environments, we observe in Figure \ref{fig:procgen} that both \textsc{Adam} PPO and CReLU encounter a continually degrading loss of plasticity as these distribution shifts are introduced. In contrast, \textsc{Trac} PPO avoids this loss of plasticity, which contributes to its rapid reward increase when adapting to new levels.  In the cumulative reward across all the Procgen levels, \textsc{Trac} PPO reveals normalized average improvements of 3,212.42\% and 120.88\% over \textsc{Adam} PPO and CReLU respectively (see Table \ref{tab:cum_reward}). For later levels, in all games, \textsc{Trac} PPO's reward does not decline as sharply as the baselines, potentially indicating positive transfer of skills from one level to the next.

One key advantage of \textsc{Trac} is that it functions as an optimizer, making it orthogonal to various policy methods such as PPO, as well as other baselines like Online EWC \citep{schwarz2018progress}, IMPALA \citep{espeholt2018impala}, Modulating Masks \citep{ben2023lifelong}, and CLEAR \citep{rolnick2019experience}. In Appendix \ref{subsection:lrl_baselines}, we evaluate these methods using both \textsc{Trac} and \textsc{Adam} on the Procgen setup. We find that in every environment, \textsc{Trac} improves the performance of these algorithms.

\textbf{Atari}\quad The Arcade Learning Environment (ALE) Atari 2600 benchmark is a collection of classic arcade games designed to assess reinforcement learning agents' performance across a range of diverse gaming scenarios \citep{DBLP:journals/corr/abs-1207-4708}. We introduce distribution shifts by switching to a new Atari game every 4 million timesteps, where each game switch introduces a new task. This benchmark is more challenging compared to OpenAI Procgen: it requires the agent to handle distribution shifts in both the input (state) and the target (reward).

\begin{figure}
    \centering
    \includegraphics[width=1\textwidth]{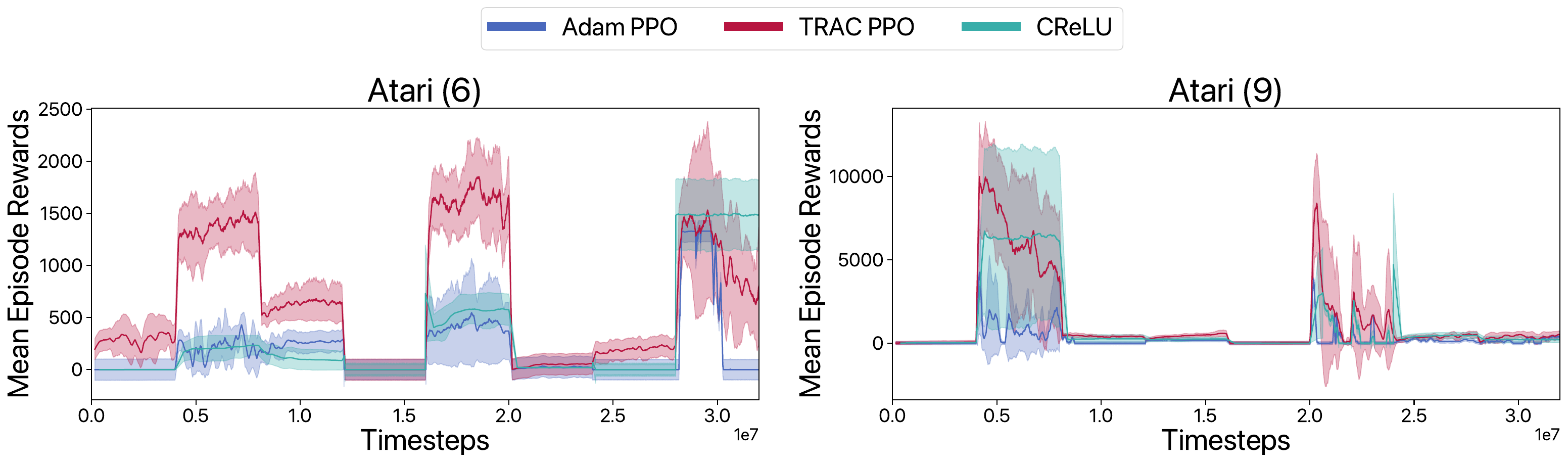}
    \caption{Reward in the lifelong Atari environments, across games with action spaces of 6 and 9.  \textsc{Trac} PPO rapidly adapts to new tasks, in contrast to the \textsc{Adam} PPO and CReLU which struggle to achieve high reward, indicating mild loss of plasticity.}
    \label{fig:atari}
    \vspace{-10pt}
\end{figure}

In this experiment, we assessed two online settings distinguished by games with action spaces of 6 and 9. From Figure \ref{fig:atari}, both \textsc{Adam} PPO and CReLU sometimes failed to learn in certain games. In contrast, \textsc{Trac} PPO shows a substantial increase in reward over different games compared to the baselines. For example, during the first 12 million steps (3 games) in Atari 6, \textsc{Trac} PPO not only achieves a significantly higher mean reward but also demonstrates rapid reward increase. Over both experiment settings, \textsc{Trac} PPO shows an average normalized improvement of 329.73\% over \textsc{Adam} PPO and 68.71\% over CReLU (Table \ref{tab:cum_reward}). In rare instances, such as the last 2 million steps of Atari 6, CReLU performs comparably to \textsc{Trac} PPO. This observation aligns with findings from \citep{pmlr-v232-abbas23a}, which noted that CReLU tends to avoid plasticity loss in continual Atari setups.

\textbf{Gym Control}\quad We use the CartPole-v1 and  Acrobot-v1 environments from the Gym Classic Control suite, along with LunarLander-v2 from Box2d Control. To introduce distribution shifts, \cite{mendez2020lifelong} periodically alters the environment dynamics. Although such distribution shifts pose only mild challenges for robust methods like PPO with \textsc{Adam} (Appendix~\ref{section:gravity-cartpole}). We instead implement a more challenging form of distribution shift. Every 200 steps we perturb each observation dimension with random noise within a range of $\pm 2$, treating each perturbation phase as a distinct task.

\definecolor{mycolor}{RGB}{26, 188, 156}
\colorlet{highlight}{mycolor!30}  

\begin{table}[ht]
\centering
\caption{Cumulative sum of mean episode reward for \textsc{Trac} PPO, \textsc{Adam} PPO, and CReLU on Procgen, Atari, and Gym Control environments. Rewards are scaled by $10^5$; higher is better.}
\label{tab:cum_reward}
\begin{tabular}{lccc}
\hline
\textbf{Environment} & \textbf{\textsc{Adam} PPO} & \textbf{CReLU} & \textbf{\textcolor{black}{\textsc{Trac} PPO (Ours)}} \\
\hline
Starpilot & $3.4$ & $3.6$ & \cellcolor{highlight}\textcolor{black}{$\mathbf{12.5}$} \\
Dodgeball & $1.9$ & $2.3$ & \cellcolor{highlight}\textcolor{black}{$\mathbf{5.2}$} \\
Chaser & $1.4$ & $1.7$ & \cellcolor{highlight}\textcolor{black}{$\mathbf{2.2}$} \\
Fruitbot & $0.1$ & $1.0$ & \cellcolor{highlight}\textcolor{black}{$\mathbf{1.8}$} \\
CartPole & $5.1$ & $1.2$ & \cellcolor{highlight}\textcolor{black}{$\mathbf{39.6}$} \\
Acrobot & $-14.3$ & $-13.9$ & \cellcolor{highlight}\textcolor{black}{$\mathbf{-12.9}$} \\
LunarLander & $-21.7$ & $-19.4$ & \cellcolor{highlight}\textcolor{black}{$\mathbf{-8.6}$} \\
Atari 6 & $3.1$ & $4.8$ & \cellcolor{highlight}\textcolor{black}{$\mathbf{10.5}$} \\
Atari 9 & $3.9$ & $17.0$ & \cellcolor{highlight}\textcolor{black}{$\mathbf{20.2}$} \\
\hline
\end{tabular}
\end{table}

\begin{figure}
    \centering
    \includegraphics[width=\textwidth]{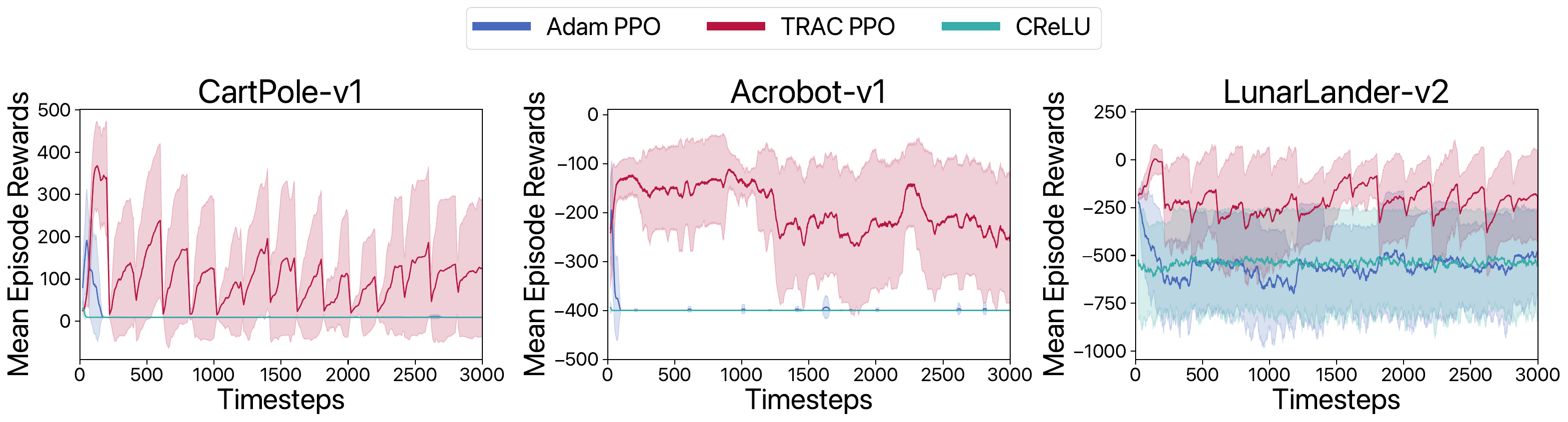}
    \caption{Reward performance across CartPole, Acrobot, and LunarLander Gym Control tasks. Both \textsc{Adam} PPO and CReLU experience extreme plasticity loss, failing to recover after the initial distribution shift. Conversely, \textsc{Trac} PPO successfully avoids such plasticity loss, rapidly adapting when facing extreme distribution shifts.}
    \label{fig:control}
\end{figure}

Here (Figure \ref{fig:control}), we notice a peculiar behavior after introducing the first distribution shift in both \textsc{Adam} PPO and CReLU: policy collapse. We describe this as an \emph{extreme} form of loss of plasticity. Surprisingly, \textsc{Trac} PPO remains resistant to these extreme distribution shifts. As we see in the Acrobot experiment, \textsc{Trac} PPO shows minimal to no policy damage after the first few distribution shifts, whereas \textsc{Adam} PPO and CReLU are unable to recover a policy at all. We investigate if \textsc{Trac}'s behavior here indicates positive transfer in Appendix \ref{section:positivetransfer}.  Across the three Gym Control environments, \textsc{Trac} PPO shows an average normalized improvement of 204.18\% over \textsc{Adam} PPO and 1044.24\% over CReLU (Table \ref{tab:cum_reward}).

%% file: sections/discussion.tex
\section{Discussion}\label{section:discussion}

\textbf{Related work}\quad Combating loss of plasticity has been studied extensively in lifelong RL. A typical challenge for existing solutions is the tuning of their hyperparameters, which requires prior knowledge on the nature of the distribution shift, e.g.,  \citep{asadi2023resetting, ben2023lifelong, nikishin2023deep,sokar2023dormant,mesbahi2024tuning}. An architectural modification called CReLU is studied in \citep{pmlr-v232-abbas23a}, but our experiments suggest that its benefit might be specific to the Atari setup. Besides, \cite{pmlr-v80-abel18a,pmlr-v80-abel18b} presented a theoretical analysis of skill transfer in lifelong RL, based on value iteration. Moreover, related contributions in nonstationary RL, where reward and state transition functions also change unpredictably, are limited to theoretical sequential decision-making settings with a focus on establishing complexity bounds \citep{roy2019multipoint, cheung2020reinforcement,wei2021nonstationary, mao2022modelfree}.

Our algorithm \textsc{Trac} builds on a long line of works on parameter-free OCO (see Section~\ref{section:method}). To our knowledge, the only existing work applying parameter-free OCO to RL is \citep{jacobsen2021parameter}, which focuses on estimating the value function (i.e., policy evaluation). Our scope is different, focusing on empirical RL in lifelong problems by exploring the key connection between parameter-free OCO and regularization.

Particularly, we are inspired by the \textsc{Mechanic} algorithm from \citep{cutkosky2023mechanic}, which goes beyond the traditional convex setting of parameter-free OCO to handle stationary deep learning optimization tasks. Lifelong reinforcement learning, however, introduces a layer of complexity with its inherent nonstationarity. Furthermore, compared to \textsc{Mechanic}, \textsc{Trac} improves the scale tuner there (which is based on the \emph{coin-betting} framework; \citealp{orabona2016coin}) by the $\erfi$ algorithm that enjoys a better OCO performance guarantee. As an ablation study, we empirically compare \textsc{Trac} and \textsc{Mechanic} in the Appendix \ref{section:other_parameter-free_methods} (Table~\ref{table:trac_mechanic_comparison}). We find that \textsc{Trac} is slightly better, but both algorithms can mitigate the loss of plasticity, suggesting the effectiveness of the general ``parameter-free'' principle in lifelong RL. 

\textbf{\textsc{Trac} encourages positive transfer}\quad
In our experiments, we observe that \textsc{Trac}'s reward decline due to distribution shifts is less severe than that of baseline methods. These results may suggest \textsc{Trac} facilitates positive transfer between related tasks. To investigate this further, we compared \textsc{Trac} to a privileged weight-resetting approach, where the network's parameters are reset for each new task, in the Gym Control environments (see Appendix \ref{section:positivetransfer}). Our results show that \textsc{Trac} maintains higher rewards during tasks than privileged weight-resetting and avoids declining to the same low reward levels as privileged weight-resetting at the start of a new task (Figure \ref{fig:positivetransfer}).

\textbf{On the choice of $\theta_\refrm$}\quad In general, the reference point $\theta_{\text{ref}}$ should be good or ``safe'' for \textsc{Trac} to perform effectively. One might presume that achieving this requires ``warmstarting'', or pre-training using the underlying \textsc{Base} optimizer. While our experiments validate that such warmstarting is indeed beneficial (Appendix~\ref{section:warmstarting}), our main experiments show that even a random initialization of the policy's weight serves as a \textit{good enough} $\theta_{\text{ref}}$, even when tasks are similar (Figure \ref{fig:procgen}). 
 
This observation aligns with discussions by \cite{lyle2023understanding},  \cite{sokar2023dormant}, and \cite{pmlr-v232-abbas23a}, who suggested that persistent gradient steps away from a random initialization can deactivate ReLU activations, leading to activation collapse and loss of plasticity in neural networks. Our results also support \cite{kumar2023maintaining}'s argument that maintaining some weights close to their initial values not only prevents dead ReLU units but also allows quick adaptation to new distribution shifts. 

\textbf{Tuning $L_2$ regularization}\quad The success of \textsc{Trac} suggests that an adaptive form of regularization---anchoring to the reference point $\theta_\refrm$---may suffice to counteract both mild and extreme forms of loss of plasticity. From this angle, we further elaborate the limitation of the $L_2$ regularization approach considered in \citep{kumar2023maintaining}. It requires selecting a regularization strength parameter \(\lambda\) through cross-validation, which is incompatible with the one-shot nature of lifelong learning settings. Furthermore, it is nontrivial to select the search grid: for example, we tried the $\lambda$-grid suggested by \citep{kumar2023maintaining}, and there is no effective $\lambda$ value within the grid for the lifelong RL environments we consider. All the values are too small. 

\begin{figure}
    \centering
    \includegraphics[width=1.0\textwidth]{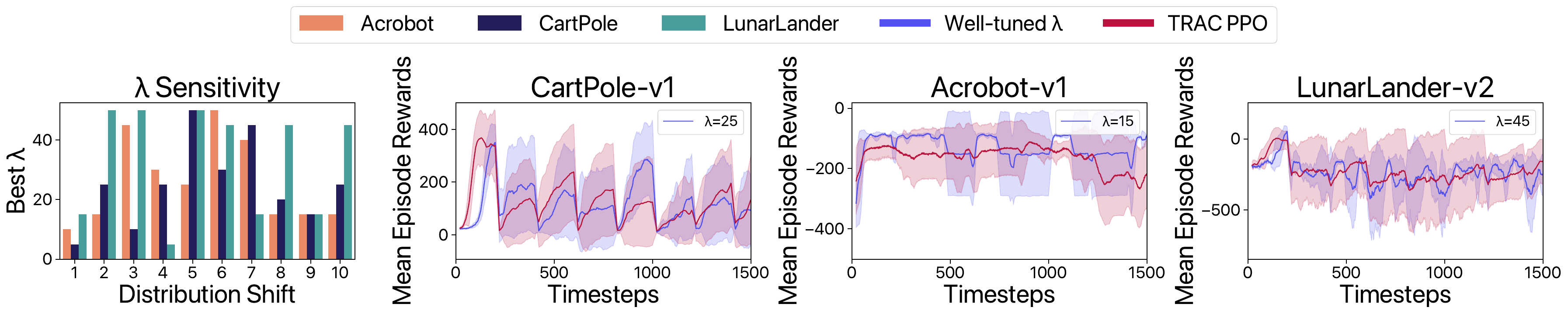}
    \caption{For each Gym Control environment and the initial ten tasks, we identified the best $\lambda$, which is the regularization strength that maximizes reward for each task's specific distribution shift. We also determined the best overall (well-tuned) $\lambda$ for each environment. The results demonstrate that each environment and each task's distribution shift is sensitive to different $\lambda$ and that \textsc{Trac} PPO performs competitively with each environment's well-tuned $\lambda$.}
    \label{fig:l2init}
    \vspace{-0.7em}
\end{figure}

Continuing this reasoning, we conduct a hyperparameter search for $\lambda$, over various larger values $[0.2, 0.8, 1, 5, 10, 15, 20, 25, 30, 35, 40, 45, 50]$. Given the expense of such experiments, only the more sample-efficient control environments are considered. We discover that each environment and task responds uniquely to these regularization strengths (see bar plot of \( \lambda \) values in Figure \ref{fig:l2init}). This highlights the challenges of tuning \( \lambda \) in a lifelong learning context, where adjusting for each environment, let alone each distribution shift, would require extensive pre-experimental analysis.

In contrast, \textsc{Trac} offers a parameter-free solution that adapts dynamically with the data in an online manner. The scaling output of \textsc{Trac} adjusts autonomously to the ongoing conditions, consistently competing with well-tuned \( \lambda \) values in the various environments, as demonstrated in the reward plots for CartPole, Acrobot, and LunarLander (Figure \ref{fig:l2init}).

\textbf{\textsc{Trac} compared to other plasticity methods}\quad Both layer normalization and plasticity injection \cite{nikishin2023deep, lyle2023understanding} have been shown to combat plasticity loss. For instance, Appendix \ref{subsection:other_plasticity_exp} Figure \ref{fig:plasticity_methods} demonstrates that both layer normalization and plasticity injection are effective at reducing plasticity loss when applied to the CartPole environment using \textsc{Adam} as a baseline optimizer. We implemented plasticity injection following the methodology laid out by \cite{nikishin2023deep}, where plasticity is injected at the start of every distribution shift. While this approach does help in reducing the decline in performance due to plasticity loss, our results indicate that it is consistently outperformed by \textsc{TRAC} across all three control environments—CartPole, Acrobot, and LunarLander. Moreover, while layer normalization improves \textsc{Adam}'s performance, it too is outperformed by \textsc{TRAC} across the same control settings (Figure \ref{fig:plasticity_methods}). Notably, combining layer normalization with \textsc{TRAC} resulted in the best performance gains.

\textbf{Near convexity of lifelong RL}\quad Our results demonstrate the rapid adaptation of \textsc{Trac}, in lifelong RL problems with complicated function approximation. From the perspective of optimization, the latter requires tackling both nonconvexity and nonstationarity, which is typically regarded intractable in theory. Perhaps surprisingly, when approaching this complex problem using the theoretical insights from OCO, we observe compelling results. This suggests a certain ``hidden convexity'' in this problem, which could be an exciting direction for both theoretical and empirical research (e.g., policy gradient methods provably converge to global optimizers in linear quadratic control~\citep{hu2023toward}). 

\textbf{Limitations}\quad While \textsc{Trac} offers robust adaptability in nonstationary environments, it can exhibit suboptimal performance at the outset. In the early stages of deployment, \textsc{Trac} might underperform compared to the baseline optimizer. We address this by proposing a warmstarting solution detailed in Appendix~\ref{section:warmstarting}, which helps increase the initial performance gap.

%% file: sections/conclusion.tex

\section{Conclusion}\label{section:conclusion}

In this work, we introduced \textsc{Trac}, a parameter-free optimizer for lifelong RL that leverages the principles of OCO. Our approach dynamically refines regularization in a data-dependent manner, eliminating the need for hyperparameter tuning. Through extensive experimentation in Procgen, Atari, and Gym Control environments, we demonstrated that \textsc{Trac} effectively mitigates loss of plasticity and rapidly adapts to new distribution shifts, where baseline methods fail.  \textsc{Trac}'s success leads to a compelling takeaway: empirical lifelong RL scenarios may exhibit more convex properties than previously appreciated, and might inherently benefit from parameter-free OCO approaches.

%% file: sections/acknowledgments.tex

\section{Acknowledgments}\label{section:acknowledgments}

We thank Ashok Cutkosky for insightful discussions on online optimization in nonstationary settings. We are grateful to David Abel for his thoughtful insights on loss of plasticity in relation to lifelong reinforcement learning. We appreciate Kaiqing Zhang and Yang Hu for their comments on theoretical and nonstationary RL. This project is partially funded by Harvard University Dean's Competitive Fund for Promising Scholarship.

%% file: sections/appendix.tex
\section{\textsc{Trac} Encourages Positive Transfer}
\label{section:positivetransfer}

\begin{figure}[h]
    \centering
    \includegraphics[width=1.0\textwidth]{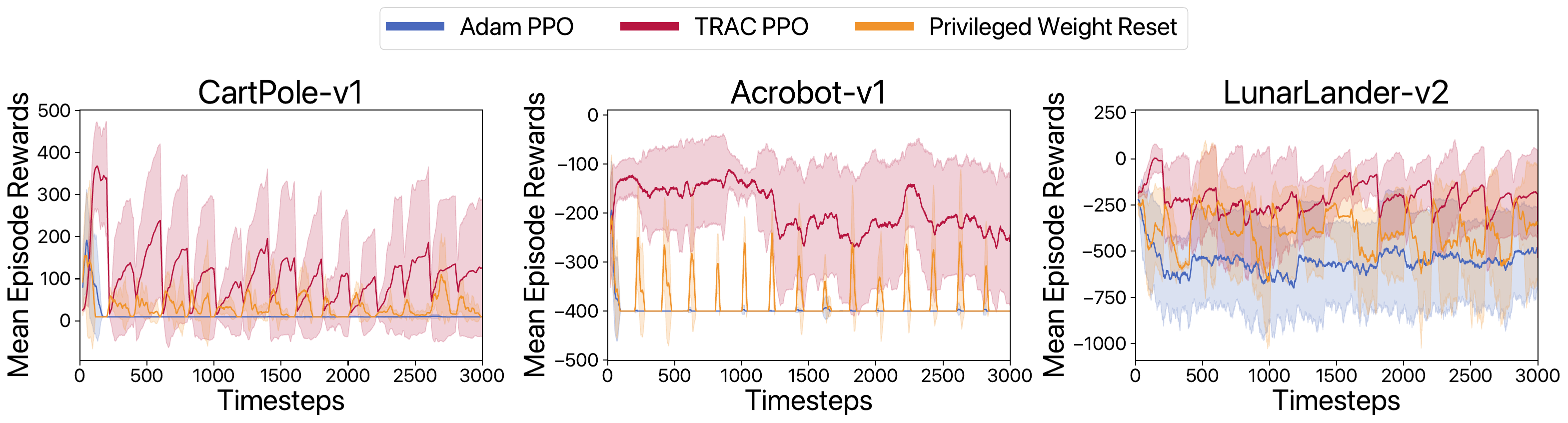}
    \caption{Reward comparison of \textsc{Trac} PPO, \textsc{Adam} PPO, and privileged weight-resetting on Cartpole-v1, Acrobot-v1, and LunarLander-v2. \textsc{Trac} PPO encourages positive transfer between tasks.}
    \label{fig:positivetransfer}
\end{figure}

To explore whether \textsc{Trac} encourages positive transfer, we introduce a privileged weight-reset baseline. This baseline is "privileged" in the sense that it knows when a distribution shift is introduced and resets the parameters to a random initialization at the start of each new task. We applied this baseline to three Gym control tasks: CartPole-v1, Acrobot-v1, and LunarLander-v2, and compared it to \textsc{Trac} PPO and \textsc{Adam} PPO, as shown in Figure \ref{fig:positivetransfer}.

We observe that the privileged weight-reset baseline exhibits spikes in reward at the beginning of each new task. Surprisingly, \textsc{Trac} maintains even higher rewards than the privileged weight-reset baseline, even at its peak learning phases. Additionally, \textsc{Trac}'s reward does not decline to the reward seen at the start of new tasks with privileged weight-resetting (\textsc{Trac} does not have to "start over" with each task), suggesting that \textsc{Trac} successfully transfers skills positively between tasks.
\section{Warmstarting}
\label{section:warmstarting}
In our theoretical framework, we hypothesize that a robust parameter initialization, denoted as $\theta_{\refrm}$, could enhance the performance of our models, suggesting that empirical implementations might benefit from initializing parameters using a base optimizer such as \textsc{Adam} prior to deploying \textsc{Trac}. Contrary to this assumption, our experimental results detailed in Section \ref{section:experiment} reveal that warmstarting is not essential for \textsc{Trac}'s success. Below, we examine the performance of \textsc{Adam} PPO and \textsc{Trac} PPO when warmstarted.

\begin{figure}[ht]
    \centering
    \includegraphics[width=1.0\textwidth]{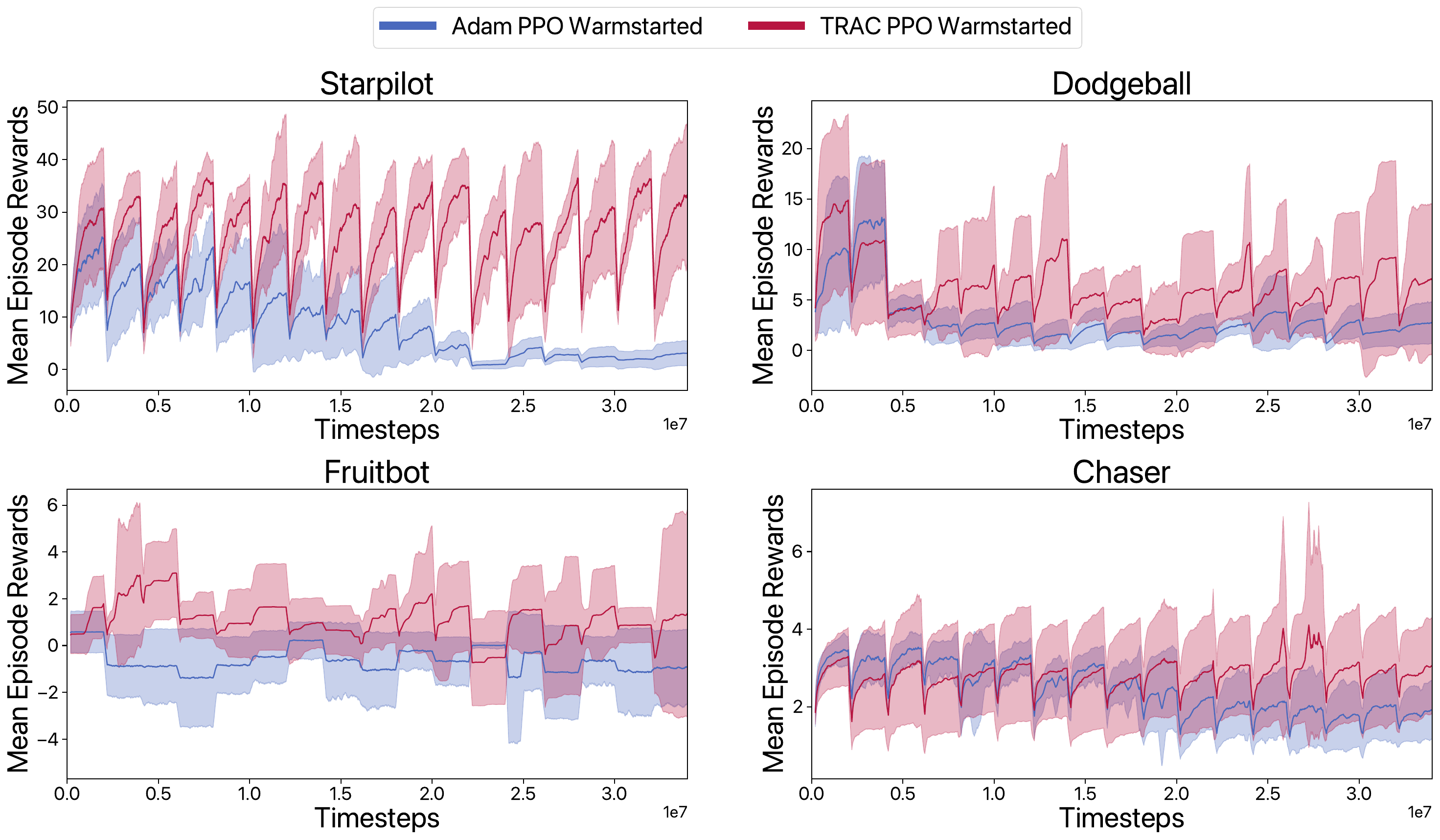}
    \caption{Reward in the lifelong Procgen environments for StarPilot, Dodgeball, Fruitbot, and Chaser with warmstarted \textsc{Trac} PPO and warmstarted \textsc{Adam} PPO. Inital performance of \textsc{Trac} PPO is improved with warmstarting and continues to avoid loss of plasticity. 
    }
    \label{fig:procgen_warm}
\end{figure}

Both \textsc{Trac} PPO and \textsc{Adam} PPO were warmstarted using \textsc{Adam} for the initial 150,000 steps in all games for the Atari and Procgen environments, and for the first 30 steps in the Gym Control experiments. As seen in Figure \ref{fig:procgen_warm}, in games like Starpilot, Fruitbot, and Dodgeball, \textsc{Trac} PPO surpasses \textsc{Adam} PPO in the first level/task of the online setup, with its performance closely matching that of \textsc{Adam} PPO in Chaser. Importantly, \textsc{Trac} PPO continues to avoid the loss of plasticity encountered by \textsc{Adam} PPO, even when both are warmstarted. This makes sense since all of the distributions share some foundational game dynamics; the initial learning phases likely explore these dynamics, so leveraging a good parameter initialization to regularize in this early region can be beneficial for \textsc{Trac}—we observe that forward transfer occurs somewhat in later level distribution shifts as the reward does not drop back to zero where it initially started from. 

Our findings indicate that warmstarting does not confer a significant advantage in the Atari games. This makes sense because a parameter initialization that is good in one game setting is likely a random parameterization for another setting, which is equivalent to the setup without warmstarting where \textsc{Trac} regularizes towards a random parameter initialization. In the Gym Control experiments although warmstarted \textsc{Trac} PPO manages to avoid the extreme plasticity loss and policy collapse seen in warmstarted \textsc{Adam} PPO, it does not perform as well as non-warmstarted \textsc{Trac} PPO. This result underscores that the efficacy of warmstarting is environment-specific and highlights the challenge in predicting when \textsc{Adam} PPO may achieve a parameter initialization that is advantageous for \textsc{Trac} PPO to regularize towards.

From an overall perspective, warmstarting \textsc{Trac} PPO in every setting still shows substantial improvement over \textsc{Adam} PPO (Table \ref{tab:cum_reward_warmstart}).
\begin{figure}[ht]
    \centering
    \includegraphics[width=\textwidth]{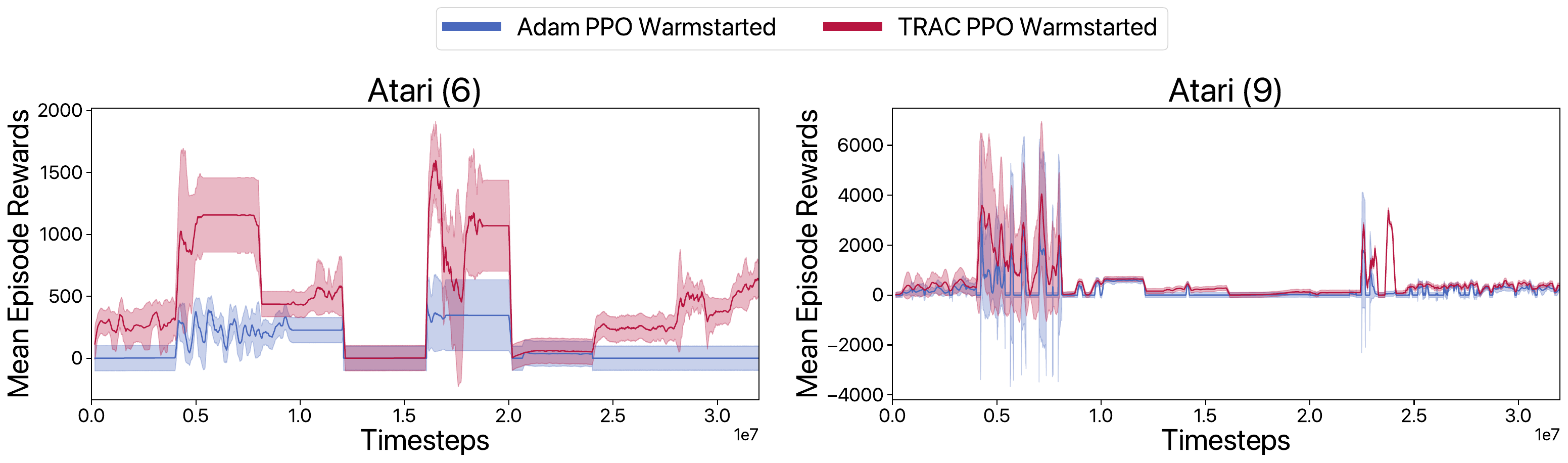}
    \caption{Reward in the lifelong Atari environments with warmstarted \textsc{Trac} PPO and warmstarted \textsc{Adam} PPO. No significant benefit is found by warmstarting \textsc{Trac} PPO compared to not warmstarting it.}
    \label{fig:atari_warm}
\end{figure}
\begin{figure}[ht]
    \centering
   \includegraphics[width=\textwidth]{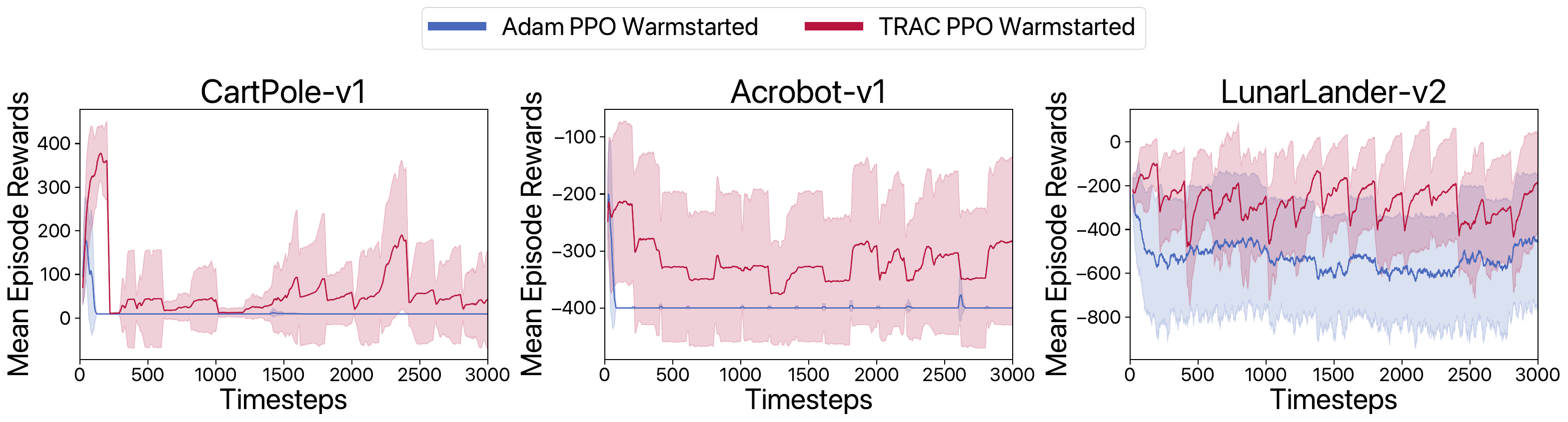}
    \caption{Reward in the lifelong Gym Control environments for CartPole-v1, Acrobot-v1, and LunarLander-v2 with warmstarted \textsc{Trac} PPO and warmstarted \textsc{Adam} PPO. \textsc{Trac} PPO still avoids loss of plasticity and policy collapse.}
    \label{fig:control_warm}
\end{figure}
\begin{table}[!ht]
\centering
\caption{Cumulative sum of mean episode reward over all distributions for \textsc{Adam} PPO warmstarted and \textsc{Trac} PPO warmstarted on Procgen, Atari, and Gym Control environments. Rewards are scaled by $10^5$; higher is better.}
\label{tab:cum_reward_warmstart}
\begin{tabular}{lcc} 
\hline
\textbf{Environment} & \textbf{\textsc{Adam} PPO} & \textbf{\textcolor{black}{\textsc{Trac} PPO (Ours)}} \\
\hline
Starpilot & $3.0$ & \cellcolor{highlight}\textcolor{black}{$\mathbf{10.2}$} \\
Dodgeball & $1.2$ & \cellcolor{highlight}\textcolor{black}{$\mathbf{2.5}$} \\
Chaser & $1.3$ & \cellcolor{highlight}\textcolor{black}{$\mathbf{1.6}$} \\
Fruitbot & $-0.4$ & \cellcolor{highlight}\textcolor{black}{$\mathbf{0.6}$} \\
CartPole & $4.6$ & \cellcolor{highlight}\textcolor{black}{$\mathbf{22.8}$} \\
Acrobot & $-142.9$ & \cellcolor{highlight}\textcolor{black}{$\mathbf{-114.5}$} \\
LunarLander & $-190.7$ & \cellcolor{highlight}\textcolor{black}{$\mathbf{-97.3}$} \\
Atari6 & $16.7$ & \cellcolor{highlight}\textcolor{black}{$\mathbf{72.2}$} \\
Atari9 & $34.6$ & \cellcolor{highlight}\textcolor{black}{$\mathbf{80.6}$} \\
\hline
\end{tabular}
\end{table}

\section{Other RL Baselines}
\label{subsection:lrl_baselines}
While PPO is a widely used policy gradient method in reinforcement learning, it is not the only approach applicable to lifelong RL. Other continual RL methods, such as IMPALA \citep{espeholt2018impala}, Online EWC \citep{schwarz2018progress}, CLEAR \citep{rolnick2019experience}, and Modulating Masks \citep{ben2023lifelong}, are designed to address challenges like catastrophic forgetting in dynamic, nonstationary environments. We incorporated these algorithm implementations adapted from the code from \cite{ben2023lifelong} into our experiments to offer a more comprehensive evaluation. These methods vary in their mechanisms for maintaining task performance over time but may still suffer from plasticity loss in later stages of training. 

\begin{figure}[h]
    \centering
    \includegraphics[width=1\textwidth]{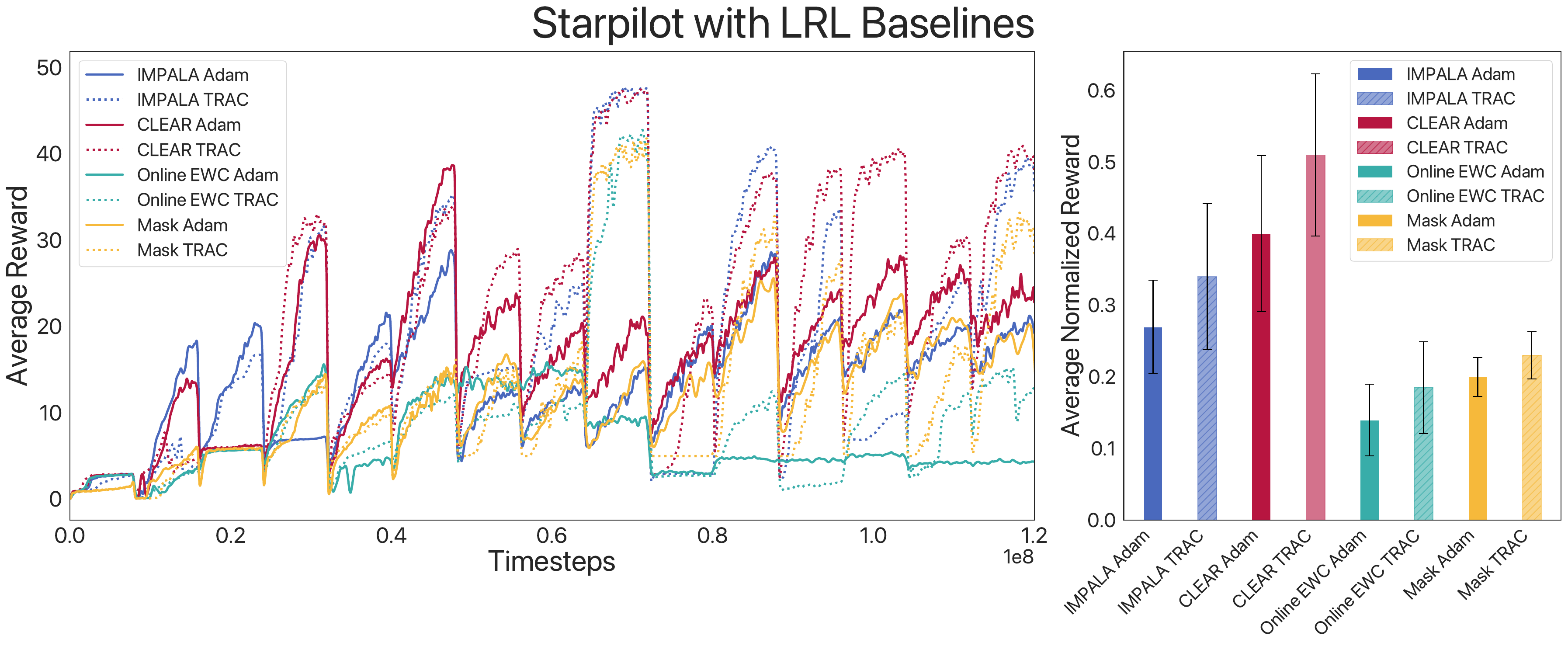}
    \caption{Performance comparison between Adam-based and \textsc{Trac}-based continual RL methods (IMPALA, Online EWC, CLEAR, Modulating Masks) in Starpilot. While \textsc{Adam} suffers from plasticity loss in later levels, \textsc{Trac} effectively mitigates this and maintains better performance over distribution shifts. For clarity, standard deviation fills are omitted here but included in the bar plot.}
    \label{fig:lrl_starpilot}
\end{figure}

\textbf{Mitigating plasticity loss across policy methods:}  
Figure \ref{fig:lrl_starpilot} demonstrates the performance of various continual RL methods when paired with \textsc{Adam} and \textsc{Trac} optimizers. The results indicate that when using \textsc{Adam}, methods like IMPALA, Online EWC, CLEAR, and Modulating Masks exhibit a noticeable decline in performance over time due to plasticity loss, particularly in later levels of the Procgen environments. In contrast, pairing these methods with \textsc{Trac} instead of \textsc{Adam} leads to significant improvements, mitigating plasticity loss and enhancing reward performance across subsequent distribution shifts.

\begin{figure}[h]
    \centering
    \includegraphics[width=1\textwidth]{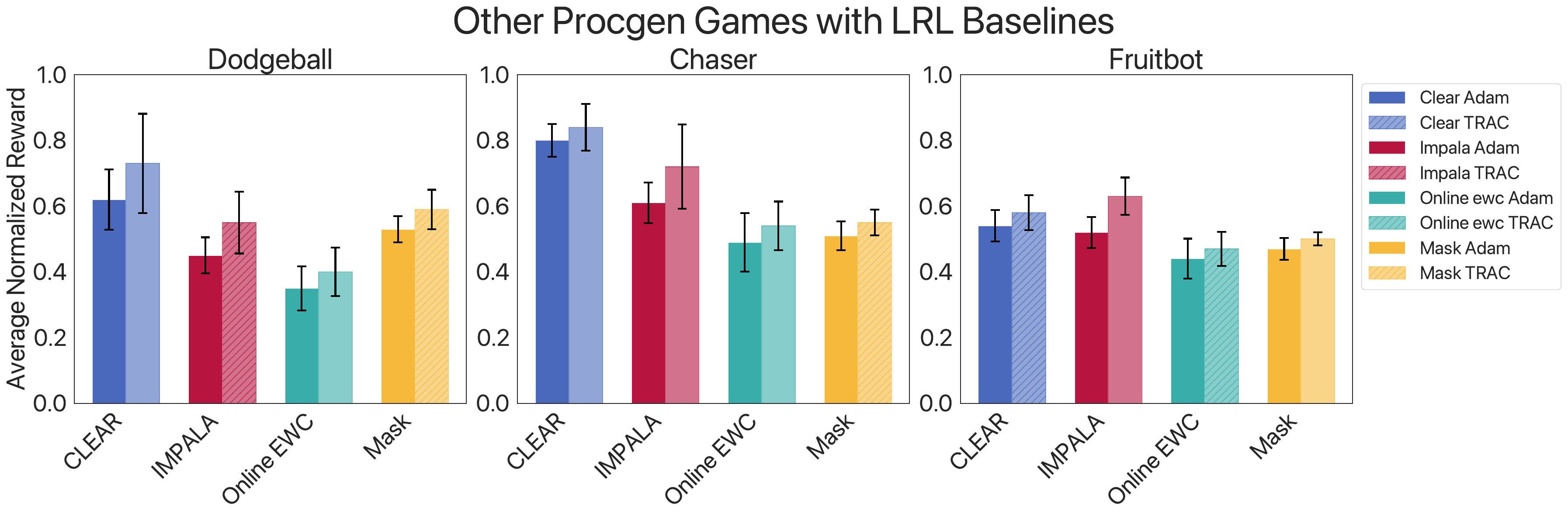}
    \caption{Average normalized rewards over five seeds and 120M timesteps for Dodeball, Chaser, and Fruitbot. Each method (IMPALA, Online EWC, CLEAR, and Modulating Masks) is evaluated using both Adam and \textsc{Trac}. \textsc{Trac} consistently outperforms \textsc{Adam} across all methods and environments.}
    \label{fig:other_lrl_procgen_normalized_rewards}
\end{figure}

To quantify these improvements, Figures \ref{fig:other_lrl_procgen_normalized_rewards}, \ref{fig:lrl_starpilot} present the average normalized rewards over five seeds and 120M timesteps for each method across four different Procgen environments: Starpilot, Dodgeball, Chaser, and Fruitbot. Across all environments, methods that use \textsc{Trac} outperform their Adam-based counterparts, consistently maintaining higher rewards over time.

On average, across the Procgen environments, \textsc{Trac} led to performance improvements over \textsc{Adam} by the following margins: \textbf{21.83\%} for IMPALA, \textbf{15.86\%} for Online EWC, \textbf{14.41\%} for CLEAR, and \textbf{10.14\%} for Modulating Masks.

\textbf{General Applicability of \textsc{Trac}:}  
It is important to highlight that \textsc{Trac} is orthogonal to the learning or policy algorithms themselves. It can be seamlessly integrated into various reinforcement learning architectures by simply replacing their optimizer (e.g., \textsc{Adam} or RMSPROP). Our results demonstrate that \textsc{Trac} enhances performance across different algorithms and environments, consistently outperforming \textsc{Adam} in mitigating plasticity loss.

\section{Gravity Based Distribution Shifts}
\label{section:gravity-cartpole}

One method to introduce distribution changes in reinforcement learning environments is by altering the dynamics \cite{mendez2020lifelong}, such as adjusting the gravity in the CartPole environment. In this set of experiments, we manipulate the gravity by a magnitude of ten, randomly adding noise for one distribution shift, and then inversely, dividing by ten and adding random noise for the next shift. This process continues throughout the experiment.

\begin{figure}[h]
    \centering
    \includegraphics[width=0.7\textwidth]{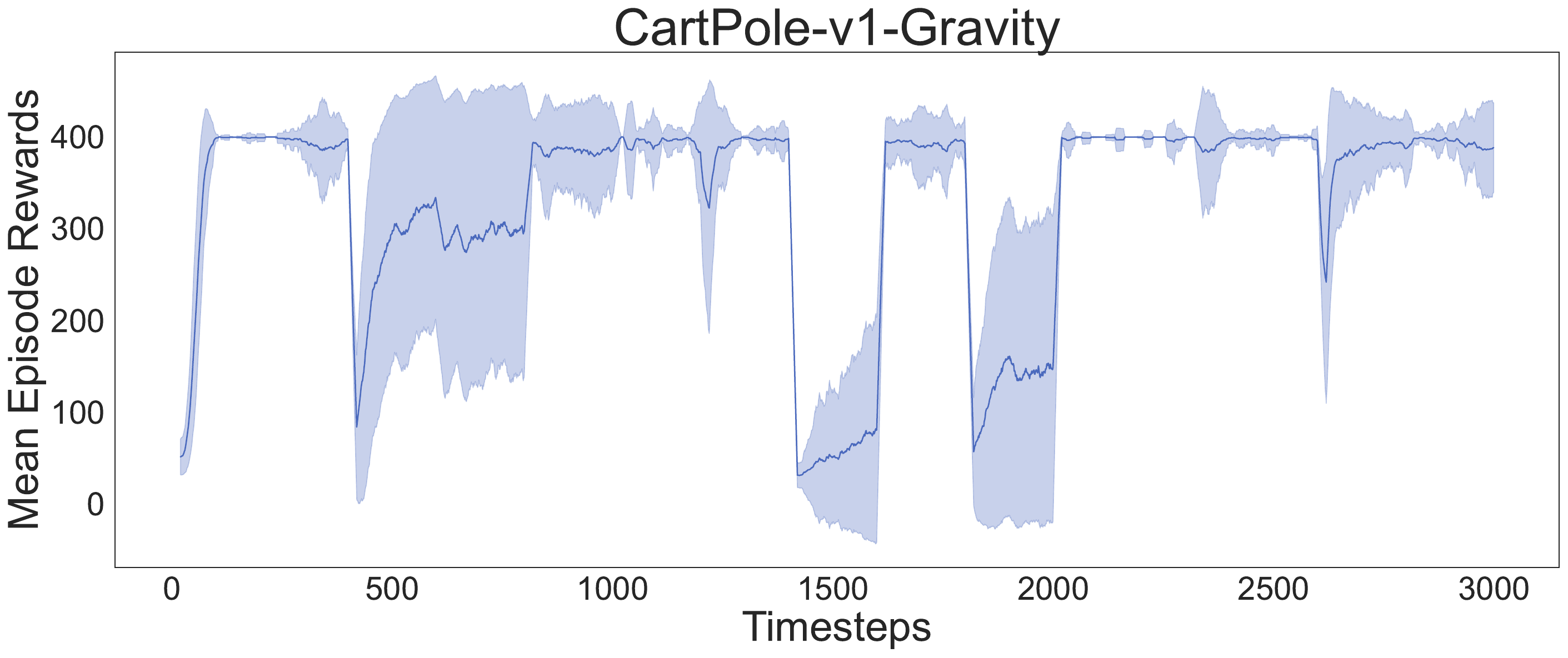}
    \caption{Mean Episode Reward for \textsc{Adam} PPO on CartPole-v1 with varying gravity. \textsc{Adam} PPO demonstrates robust policy recovery across most gravity-based distribution shifts.}
    \label{fig:cartpole_gravity}
\end{figure}

Our observations suggest that \textsc{Adam} PPO is robust to such dynamics-based distribution shifts, as shown in Figure~\ref{fig:cartpole_gravity}. This indicates that while \textsc{Adam} PPO implicitly models the dynamics of the environment well—where changes in dynamics minimally impact performance—it struggles more with adapting to out-of-distribution observations such as seen in the main experiments (Figure \ref{fig:control}) and in the warmstarting experiments (Figure \ref{fig:control_warm}).

\section{LayerNorm, Plasticity Injection, and Weight Decay}
\label{subsection:other_plasticity_exp}

To evaluate \textsc{Trac} alongside other methods that aim to mitigate plasticity loss, we compare it against LayerNorm \citep{lyle2023understanding}, Plasticity Injection \citep{nikishin2023deep}, and tuning weight decay \citep{lyle2024disentanglingcausesplasticityloss}. 

\begin{figure}[ht]
    \centering
    \includegraphics[width=0.92\textwidth]{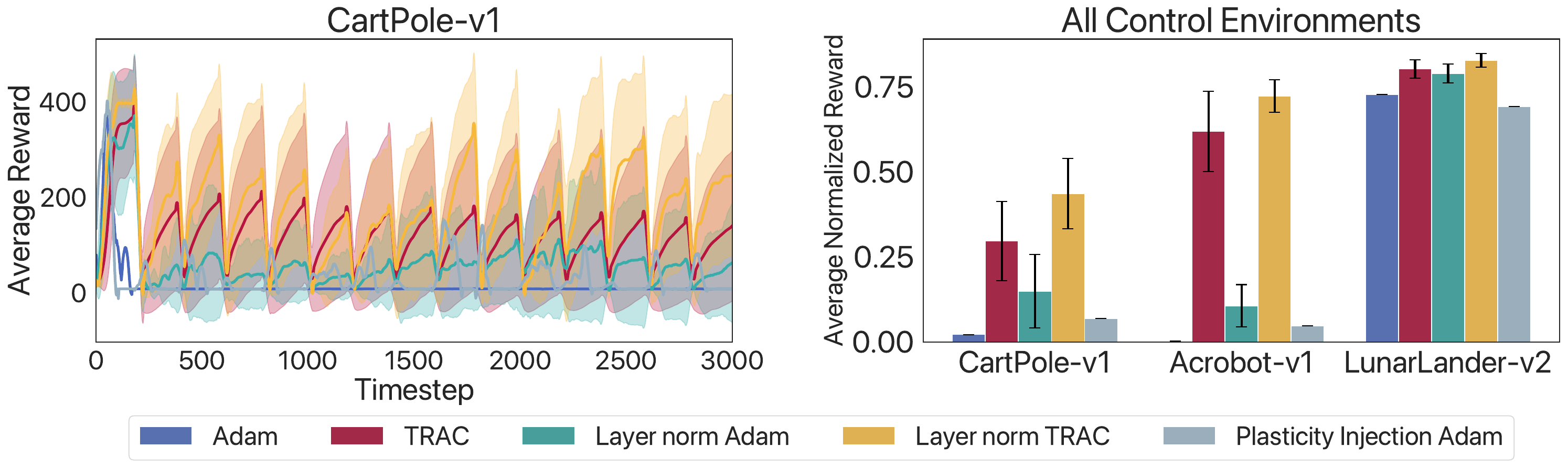}
    \caption{Performance comparison of plasticity loss mitigation techniques across Gym Control environments. Both layer normalization and plasticity injection reduce plasticity loss when applied with \textsc{Adam}. \textsc{Trac} outperforms both layer norm \textsc{Adam} and plasticity injection \textsc{Adam}, with the combination of layer norm and \textsc{Trac} achieving the highest performance.}

    \label{fig:plasticity_methods}
\end{figure}

As discussed in Section \ref{section:discussion}, we confirm that both layer normalization and plasticity injection (applied at the start of every distribution shift) \citep{nikishin2023deep, lyle2023understanding} are effective in reducing plasticity loss (Figure \ref{fig:plasticity_methods}). While these methods help slow the decline in performance due to plasticity loss, \textsc{Trac} consistently outperforms them across the three Gym Control environments. Importantly, because \textsc{Trac} is an optimizer, it can be combined with layer normalization, and doing so resulted in the best performance gains in our Control setups.

\begin{figure}[ht]
    \centering
    \includegraphics[width=1\textwidth]{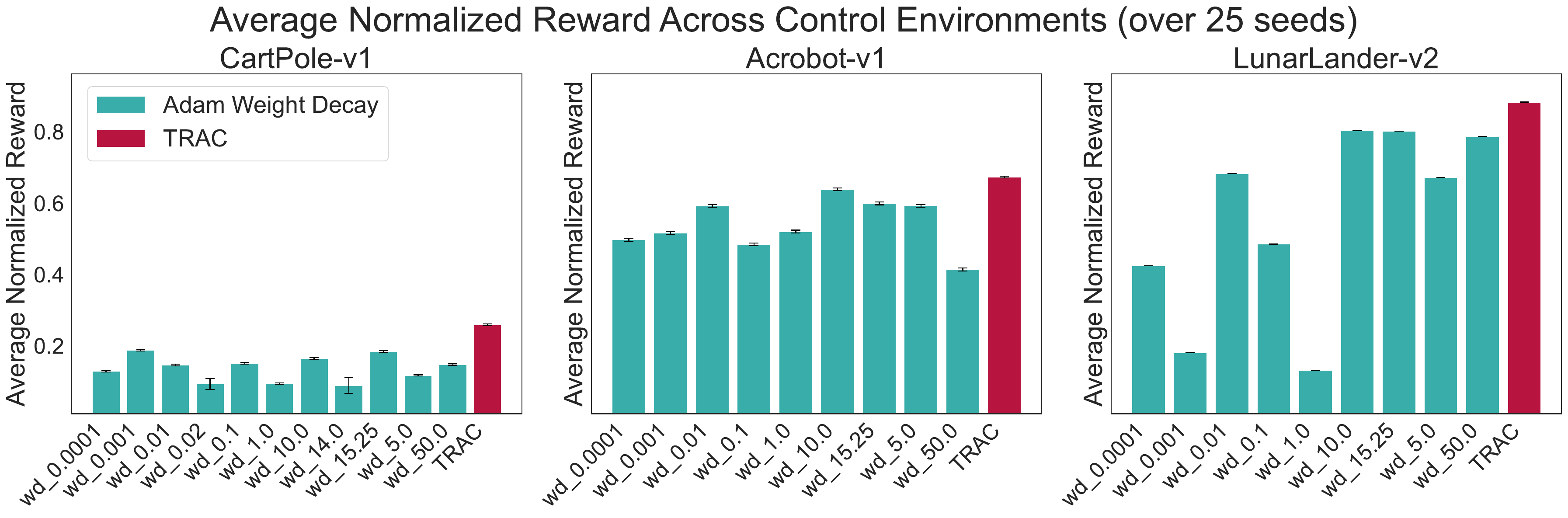}
    \caption{Effect of weight decay on performance in the three Gym Control environments. Bar plots show the average normalized rewards over 25 seeds for different weight decay values using \textsc{Adam} across 3000 timesteps, compared to \textsc{Trac} with no weight decay.}

\label{fig:weight_decay}
\end{figure}

\textbf{Tuning weight decay:}  
In addition to LayerNorm and Plasticity Injection, we also evaluated the effects of tuning weight decay using PyTorch’s AdamW optimizer. We conducted a hyperparameter sweep across three control environments with 15 seeds for each of the following weight decay values: 0.0001, 0.001, 0.01, 0.1, 1.0, 5.0, 10.0, 15.0, and 50.0. Figure \ref{fig:weight_decay} presents the average normalized reward for each weight decay value over 15 seeds and 3000 timesteps, compared to \textsc{Trac}.

The results indicate that while tuning weight decay with Adam does provide some benefit, these values consistently underperform in comparison to \textsc{Trac} across all three control environments. Figure \ref{fig:weight_decay} plots the performance of the best-performing weight decay value with Adam over 10 distribution shifts in the control environments. We observe that weight decay values are highly sensitive to the specific environment and the nature of the distribution shift. 

Interestingly, in our initial experiments, we set the weight decay to zero, yet \textsc{Trac} still outperformed Adam with various weight decay values. This suggests that while weight decay can mitigate plasticity loss to some extent, it does not match the overall effectiveness of \textsc{Trac}.

\section{Scaling-Value Convergence}
\label{subsection:scaling_value}

As discussed in the algorithm section (see Section~\ref{section:method}), \textsc{Trac} operates as a meta-algorithm on top of a standard optimizer, denoted as \textsc{Base}. The crucial component of \textsc{Trac} involves the dynamic adjustment of the scaling parameter \( S_{t+1} \), managed by the tuner algorithm (Algorithm~\ref{alg:tuner}). This parameter is data-dependent and typically ranges between $[0,1]$. The weight update \( \theta_{t+1} \) is consequently defined as a convex combination of the current optimizer's weight \( \theta_t^{\textsc{Base}} \) and a predetermined reference point \( \theta_{\text{ref}} \).

This section presents the convergence behavior of the scaling parameter \( S_{t+1} \) across different environments, analyzed through the mean values over multiple seeds. 

\begin{figure}[ht]
    \centering
    \includegraphics[width=1\textwidth]{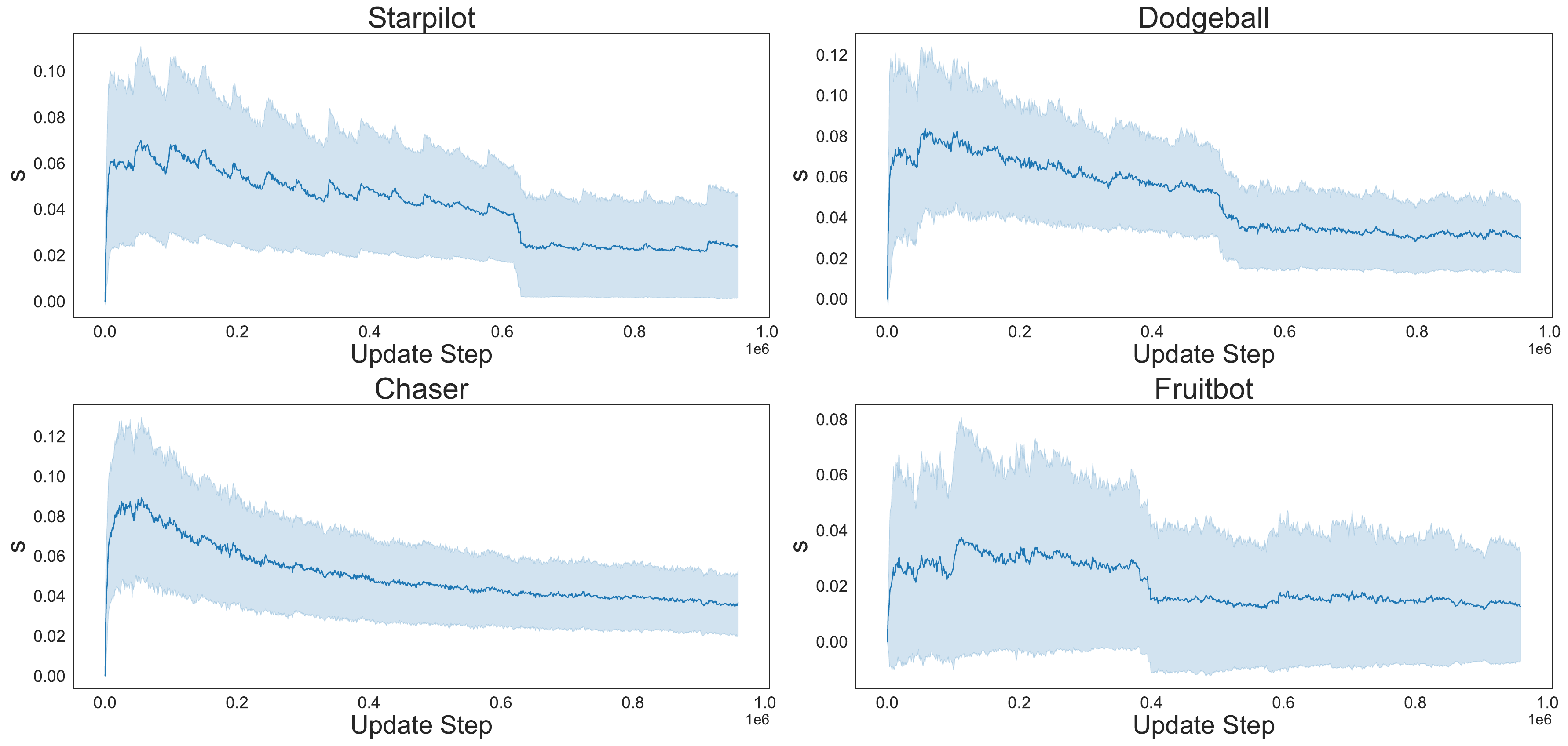}
    \caption{Convergence of the scaling parameter \( S_{t+1} \) in the Procgen environments.}
    \label{fig:scaling-procgen}
\end{figure}

\begin{figure}[ht]
    \centering
    \includegraphics[width=1\textwidth]{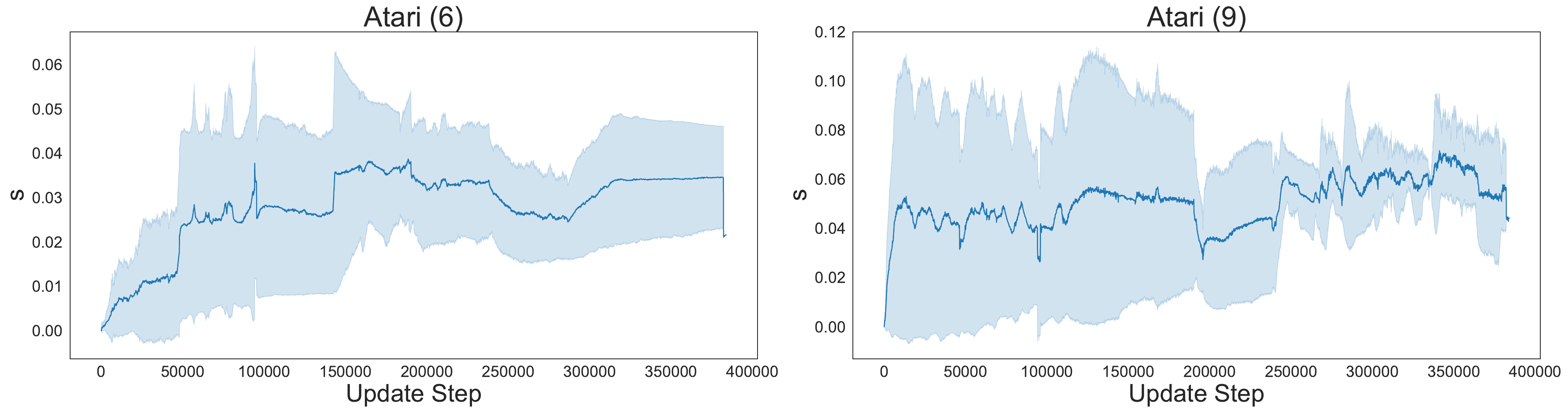}
    \caption{Evolution of the scaling parameter \( S_{t+1} \) in the Atari environments. Here we don't see a meaningful convergence of \( S_{t+1} \). }
    \label{fig:scaling-atari}
\end{figure}

\begin{figure}[ht]
    \centering
    \includegraphics[width=1\textwidth]{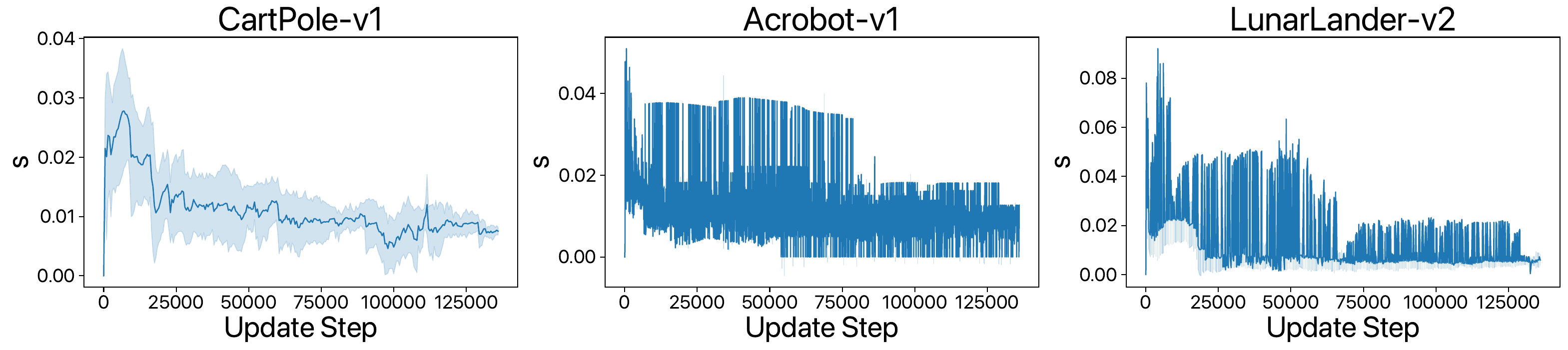}
    \caption{Convergence of the scaling parameter \( S_{t+1} \) in the Gym Control environments.}
    \label{fig:scaling-control}
\end{figure}

The convergence of the scaling parameter \( S_{t+1} \) observed across the Procgen and Gym Control environments, as depicted in Figures \ref{fig:scaling-procgen} and \ref{fig:scaling-control}, reflects a good scaling value that effectively determines the strength of regularization towards the initialization points, yielding robust empirical outcomes in lifelong RL settings. Interestingly, in Procgen environments, this converged scaling value exhibits consistency across various games, typically hovering between 0.02 and 0.03, as shown in Figure \ref{fig:scaling-procgen}. In contrast, in the Gym Control environments, the scaling values are lower, ranging between 0.005 and 0.01, as illustrated in Figure \ref{fig:scaling-control}.

\section{Comparison to \textsc{Mechanic}}
\label{section:other_parameter-free_methods}

In our analysis, we extend the examination to other OCO-based optimizers within the lifelong RL setup. Table \ref{table:trac_mechanic_comparison} presents a comparative assessment of \textsc{Trac} PPO and \textsc{Mechanic} PPO \citep{cutkosky2023mechanic} for the lifelong Gym Control tasks (with 300 seed runs).  The p-values were calculated using two-sample t-tests to test the hypothesis that the means between \textsc{Trac} and \textsc{Mechanic} are the same (Null Hypothesis, \(H_0\)) against the alternative hypothesis that they are different (Alternative Hypothesis, \(H_1\)). The results indicate that while \textsc{Mechanic} effectively mitigates plasticity loss and adapts quickly to new distribution shifts, it slightly underperforms in comparison to \textsc{Trac}. 

\begin{table}[ht]
    \centering
    \caption{Performance comparison between \textsc{Trac} and \textsc{Mechanic} across three Gym Control environments. The mean, standard error, and p-values reflect the performance over multiple runs, with bolded values highlighting \textsc{Trac}'s superior results.}
    \label{table:trac_mechanic_comparison}
    \begin{tabular}{|l|l|c|c|c|}
        \hline
        \textbf{Task} & \textbf{Method} & \textbf{Mean} & \textbf{Std Error} & \textbf{p-value} \\
        \hline
        LunarLander-v2 & \textsc{Trac} & \textbf{0.6018} & 0.0036 & 0.0000 \\
                       & Mechanic      & 0.5755         & 0.0027 & \\
        \hline
        CartPole-v1    & \textsc{Trac} & \textbf{0.3518} & 0.0244 & 0.0021 \\
                       & Mechanic      & 0.3008         & 0.0230 & \\
        \hline
        Acrobot-v1     & \textsc{Trac} & \textbf{0.7044} & 0.0221 & 0.0000 \\
                       & Mechanic      & 0.6396         & 0.0239 & \\
        \hline
    \end{tabular}
\end{table}

\section{Experimental Setup}
\label{section: exp_setup}
\paragraph{Procgen and Atari Vision backbone}
For both the Atari and Procgen experiments, the Impala architecture was used as the vision backbone. The Impala model had 3 Impala blocks, each containing a convolutional layer followed by 2 residual blocks. The output of this is flattened and connected to a fully connected layer. The impala model parameters are initialized using Xavier uniform initialization.

\paragraph{Policy and Value Networks}
Across all experiments—including Gym Control, Atari, and Procgen—the policy and value functions are implemented using a multi-layer perceptron (MLP) architecture. This architecture processes the input features into action probabilities and state value estimates. The MLP comprises several fully connected layers activated by ReLU. The output from the final layer uses a softmax activation.

\begin{table}[!ht]
\centering
\caption{PPO Parameters for Atari, Procgen, and Gym Control Experiments}
\label{tab:ppo_parameters}
\begin{tabular}{|l|c|c|c|}
\hline
\textbf{Parameter} & \textbf{Atari} & \textbf{Procgen} & \textbf{Control} \\
\hline
Steps per update & 2,000 & 1,000 & 800 (2 episodes with 400 steps) \\
\hline
Batch size & 250 & 125 & 32 \\
\hline
Epochs per update & 3 & 3 & 5 \\
\hline
Epsilon clip for PPO & 0.2 & 0.2 & 0.2 \\
\hline
Value coefficient & 0.5 & 0.5 & 0.5 \\
\hline
Entropy coefficient & 0.01 & 0.01 & 0.01 \\
\hline
Base Optimizer & \textsc{Adam} (LR: 0.001) & \textsc{Adam} (LR: 0.001) & \textsc{Adam} (LR: 0.01) \\
\hline
Architecture & Impala + MLP & Impala + MLP & MLP \\
\hline
\end{tabular}
\end{table}

\paragraph{\textsc{Trac}}
\textsc{Trac}, for all experiments, was implemented using the same experiment-specific baseline architectures and baseline optimizer. For the Procgen and Atari experiments, the base \textsc{Adam} optimizer was configured as the same as baseline, with a learning rate of 0.001, and for the Gym Control experiments, a learning rate of 0.01 was used. Both learning rates were tested for all experiments and found to have negligible differences in performance outcomes. Other than the learning rate, we use the default \textsc{Adam} parameters, including weight decay and betas, followed by the specifications outlined in the PyTorch Documentation.\footnote{\url{https://pytorch.org/docs/stable/generated/torch.optim.Adam.html}}

The setup for \textsc{Trac} included $\beta$ values for adaptive gradient adjustments: 0.9, 0.99, 0.999, 0.9999, 0.99999, and 0.999999. Both $S_t$ and $\eps$ were initially set to (\(1 \times 10^{-8}\)). Modifications were made to a PyTorch error function library, which accepts complex inputs to accommodate the necessary computations for the imaginary error function. This library can be found at Torch Erf GitHub.\footnote{\url{https://github.com/redsnic/torch_erf}}

\paragraph{Distribution Shifts}
 In the Atari experiments, game environments were switched every 4 million steps. The sequence for games with an action space of 6 included ``BasicMath'', ``Qbert'', ``SpaceInvaders'', ``UpNDown'', ``Galaxian'', ``Bowling'', ``Demonattack'', ``NameThisGame'', while games with an action space of 9 included ``LostLuggage'', ``VideoPinball'', ``BeamRider'', ``Asterix'', ``Enduro'', ``CrazyClimber'', ``MsPacman'', ``Koolaid''.

For Procgen experiments, individual game levels were sampled using a seed value as the \emph{start\_level} parameter, which was incremented sequentially to generate new levels. Each new environment was introduced every 2 million steps.

In the Gym Control experiments, each observation dimension was randomly perturbed by a value ranging from 0 to 2. This perturbation was constant for 200 timesteps, after which a new perturbation was applied, effectively switching the environmental conditions every 200 steps.

\paragraph{Statistical Significance}
The Procgen and Atari experiments were conducted with 8 seeds/runs, while the Gym Control experiments utilized 25 seeds/runs (with the exception of the Mechanic experiments in Table \ref{table:trac_mechanic_comparison} which utilized 300 seeds). The exception was in the \(L_2\) initialization experiments, which used 15 seeds/runs per regularization strength. In Figures \ref{fig:procgen}, \ref{fig:atari}, \ref{fig:control}, \ref{fig:l2init}, \ref{fig:procgen_warm}, \ref{fig:atari_warm}, \ref{fig:control_warm},
\ref{fig:plasticity_methods},
\ref{fig:lrl_starpilot},
\ref{fig:cartpole_gravity}, the plotted lines represent the mean of all of the mean episode rewards from the different seeds/runs, and the shaded error bands indicate the standard deviation of all of the mean episode rewards from the different seeds/runs.

\paragraph{Compute Resources}
For the Procgen and Atari experiments, each was allocated a single A100 GPU, typically running for 3-4 days to complete. The Gym Control experiments were conducted using dual-core CPUs, generally concluding within a few hours. In both scenarios, an allocation of 8GB of RAM was sufficient to meet the computational demands.